\documentclass[10pt,journal,compsoc]{IEEEtran}
\ifCLASSOPTIONcompsoc
  \usepackage[nocompress]{cite}
\else
  \usepackage{cite}
\fi
\ifCLASSINFOpdf
   \usepackage[pdftex]{graphicx}
\else
\fi
\usepackage{amsmath}
 \usepackage{dblfloatfix}

\usepackage{times}
\usepackage{epsfig}
\usepackage{amssymb}
\usepackage{caption}
\usepackage{media9}
\begin{document}
%
\title{FakeCatcher: Detection of Synthetic Portrait Videos using Biological Signals}
%
%
%
%

\author{Umur~Aybars~Ciftci,
        \.Ilke~Demir,
        and~Lijun~Yin,~\IEEEmembership{Senior Member,~IEEE}
\IEEEcompsocitemizethanks{\IEEEcompsocthanksitem Umur~Aybars~Ciftci and Lijun Yin are with the Department of Computer Science, Binghamton University, Binghamton, NY. 
\IEEEcompsocthanksitem Ilke~Demir is a Senior Research Scientist at Intel Corporation, CA.\protect\\ 
E-mail: \{uciftci,lijun\}@binghamton.edu, idemir@purdue.edu
}
\thanks{Manuscript received Sept. 11, 2019; revised May 12; accepted July 4, 2020.}}

%
%

\markboth{IEEE Transactions on Pattern Analysis and Machine Intelligence,~Vol.~X, No.~X, July~2020}%
{Shell \MakeLowercase{\textit{et al.}}: Bare Demo of IEEEtran.cls for Computer Society Journals}
%



\IEEEtitleabstractindextext{%
\begin{abstract}
The recent proliferation of fake portrait videos poses direct threats on society, law, and privacy\cite{ucla}. Believing the fake video of a politician, distributing fake pornographic content of celebrities, fabricating impersonated fake videos as evidence in courts are just a few real world consequences of deep fakes. We present a novel approach to detect synthetic content in portrait videos, as a preventive solution for the emerging threat of \textit{deep fakes}. In other words, we introduce a deep fake detector. We observe that detectors blindly utilizing deep learning are not effective in catching fake content, as generative models produce formidably realistic results. Our key assertion follows that biological signals hidden in portrait videos can be used as an implicit descriptor of authenticity, because they are neither spatially nor temporally preserved in fake content. To prove and exploit this assertion, we first engage several signal transformations for the pairwise separation problem, achieving 99.39\% accuracy. Second, we utilize those findings to formulate a generalized classifier for fake content, by analyzing proposed signal transformations and corresponding feature sets. Third, we generate novel signal maps and employ a CNN to improve our traditional classifier for detecting synthetic content. Lastly, we release an ``in the wild'' dataset of fake portrait videos that we collected as a part of our evaluation process. We evaluate FakeCatcher on several datasets, resulting with 96\%, 94.65\%, 91.50\%, and 91.07\% accuracies, on Face Forensics~\cite{ff}, Face Forensics++~\cite{FF++}, CelebDF~\cite{Celeb_DF_cvpr20}, and on our new Deep Fakes Dataset respectively. In addition, our approach produces a significantly superior detection rate against baselines, and does not depend on the source, generator, or properties of the fake content. We also analyze signals from various facial regions, under image distortions, with varying segment durations, from different generators, against unseen datasets, and under several dimensionality reduction techniques. 
\end{abstract}

}

\maketitle

\IEEEdisplaynontitleabstractindextext

%
\IEEEpeerreviewmaketitle

\IEEEraisesectionheading{\section{Introduction}\label{sec:introduction}}

%
%
%
%
\IEEEPARstart{A}{s} we enter into the artifical intelligence (AI) era, the technological advancements in deep learning started to revolutionize our perspective on how we solve difficult problems in computer vision, robotics, and related areas. 
In addition, the developments in generative models (i.e., \cite{GAN, dcgan, cyclegan, vid2vid}) empower machines to increase the photorealism in the generated data and mimic the world more accurately. Even though it is easy to speculate dystopian scenarios based on both analysis and synthesis approaches, the latter brought the immediate threat on information integrity by disabling our ``natural detectors'': we cannot simply look at an image to determine its authenticity.

Following the recent initiatives for democratization of AI, generative models become increasingly popular and accessible. The widespread use of generative adversarial networks (GAN) is positively impacting some technologies: it is very easy to create personalized avatars \cite{pagan}, to produce animations \cite{ganimation}, and to complete and stylize images \cite{gfc}. Unfortunately, they are also used with malicious intent. The famous deep fake Obama video warning us about deep fakes going viral \cite{msnbc} may be the most innocent example, but there are far more devastating real-world examples which impact the society by introducing inauthentic content such as fake celebrity porn \cite{deepfakereddit}, political misinformation through fake news \cite{threatblog}, and forfeiting art using AI\cite{AIart}. This lack of authenticity and increasing information obfuscation pose real threats to individuals, criminal system, and information integrity. As every technology is simultaneously built with the counter-technology to neutralize its negative effects, we believe that it is the perfect time to develop a deep fake detector to battle with deep fakes before having serious consequences. 

We can regard counterfeiting money as an analogy. Actual bills are stamped with an unknown process so that fake money can not contain that special mark. We are looking for that special mark in authentic videos, which are already stamped by physics and biology of the real world, but generative models yet to replicate in image space because they ``do not know'' how the stamp works. As an example of such special marks, the chaotic order of the real world is found as an authenticity stamp by the study of Benes et al. \cite{muller17}. As another example, Pan et al. \cite{physicsbased} defend the same idea and propose integrating physics-based models for GANs to improve the realism of the generated samples. Biological signals (such as heart rate) have already been exploited for medical~\cite{6226654} and security purposes~\cite{Rostami:2013:HAI:2541806.2516658}, therefore we wonder: \textit{Can we depend on the biological signals as our authenticity stamps? Do generative models capture biological signals? If not, can we formulate how and why they fail to do so?}

\begin{figure*}[ht]
\centering
    \includegraphics[width=0.95\textwidth]{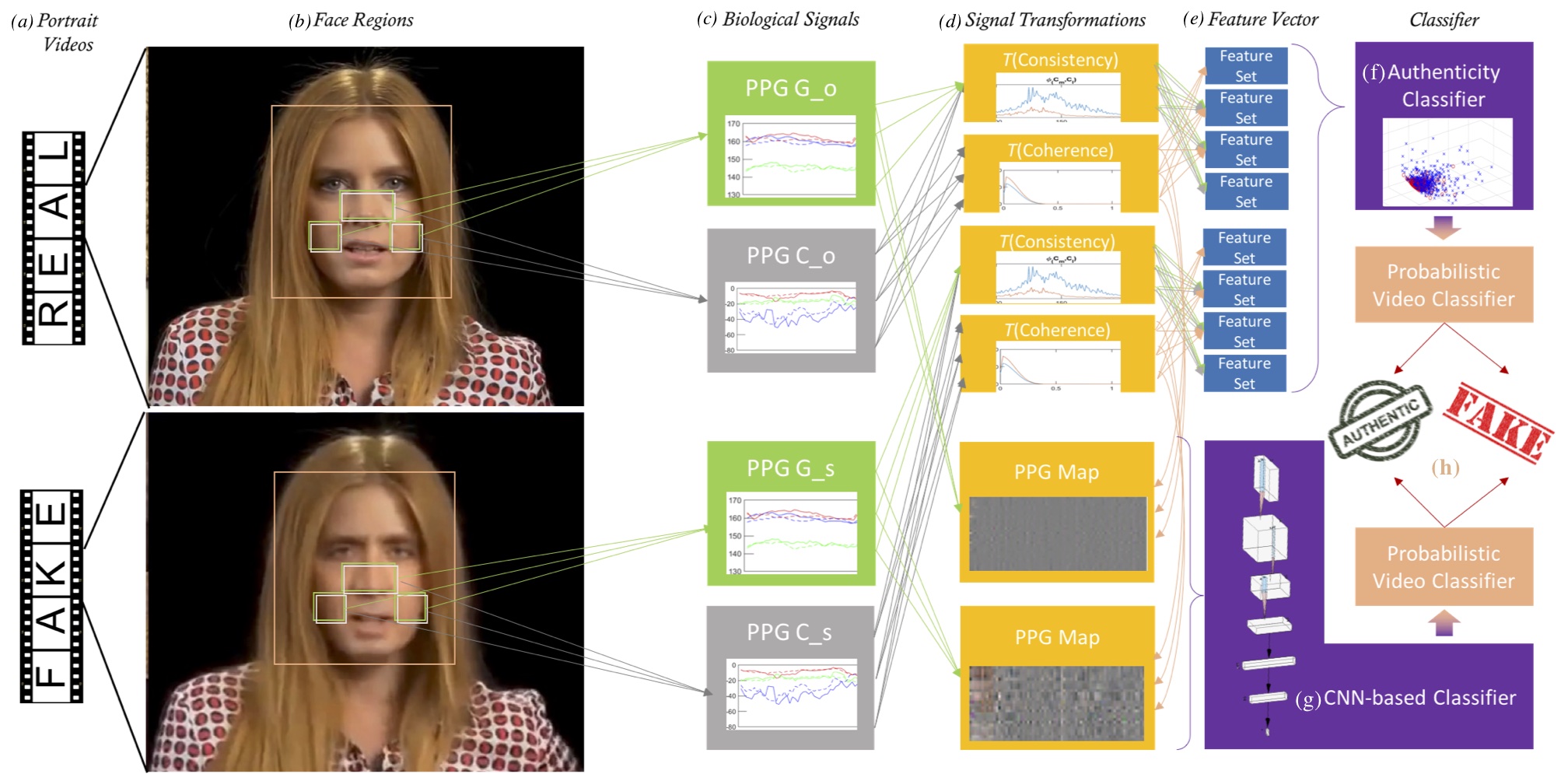}
    \caption{\textbf{Overview.} We extract biological signals (c) from face regions (b) of authentic and fake portrait video pairs (a). We apply transformations (d) to compute spatial coherence and temporal consistency, capture signal characteristics in feature sets (e) and PPG maps, and train a probabilistic SVM (f) and a CNN (g). Then, we aggregate authenticity probabilities (h) to classify the authenticity.}
  \label{fig:teaser}
  \end{figure*}

We observe that, even though GANs learn and generate photorealistic visual and geometric signals beyond the discriminative capabilities of human eyes, biological signals hidden by nature are still not easily replicable. Biological signals are also intuitive and complimentary ingredients of facial videos -- which is the domain of deep fakes. Moreover, videos, as opposed to images, contain another layer of complexity to becloud fake synthesis: the consistency in the time dimension. Together with the signals' spatial coherence, temporal consistency is the key prior to detect authentic content. Although there are recent pure deep learning approaches to detect fake content, those are limited by the specific generative model \cite{mesonet}, dataset \cite{ff}, people \cite{Agarwal_2019_CVPR_Workshops}, or hand-crafted features \cite{blink}. In contrast to all, we choose to search for some natural priors in authentic content, instead of putting some assumptions on the fake content. To complete this narrative, our approach exploits biological signals to detect fake content in portrait videos, independent of the source of creation.

Our main contributions include,
\vspace{-2mm}
\begin{itemize}
\item formulations and experimental validations of signal transformations to exploit spatial coherence and temporal consistency of biological signals, for both pairwise and general authenticity classification, 
\item a generalized and interpretable deep fake detector that operates in-the-wild,
\item a novel biological signal map construction to train neural networks for authenticity classification,
\item a diverse dataset of portrait videos to create a test bed for fake content detection in the wild.
\end{itemize}

Our system processes input videos (Figure~\ref{fig:teaser}a) by collecting fixed-length video segments with facial parts, defining regions of interests within each face (Figure~\ref{fig:teaser}b), and extracting several biological signals (Figure~\ref{fig:teaser}c) from those regions in those temporal segments. In the first part of our paper, we scrutinize the pairwise separation problem where video pairs are given but the fake one is not known. We formulate a solution by examining the extracted biological signals, their transformations to different domains (time, frequency, time-frequency), and their correlations (Figure~\ref{fig:teaser}d). In the second part, we combine the revelations from the pairwise context and interpretable feature extractors in the literature (Figure~\ref{fig:teaser}e) to develop a generalized authenticity classifier working in a high dimensional feature space (Figure~\ref{fig:teaser}f). In the third part, we transform the signals into novel signal maps of fixed-duration segments to train a simple convolutional deep fake detector network (Figure~\ref{fig:teaser}g). We also aggregate the class probabilities of segments in a video into a binary ``fake or authentic'' decision (Figure~\ref{fig:teaser}h).

To evaluate FakeCatcher, we collected over 140 online videos, totaling up to a ``Deep Fakes Dataset'' of 30GB. It is important to note that, unlike existing datasets, our dataset includes ``in the wild'' videos, independent of the generative model, resolution, compression, content, and context. Our simple convolutional neural network (CNN) achieves 91.07\% accuracy for detecting inauthentic content on our dataset, 96\% accuracy on Face Forensics dataset (FF) \cite{ff}, and 94.65\% on Face Forensics++ (FF++) \cite{FF++} dataset, outperforming all baseline architectures we compared against. We also analyzed the effects of segment durations, facial regions, face detector, image distortions, and dimensionality reduction techniques on mentioned datasets.

\section{Related Work}

Traditionally, image spoofing and forgery has been an important topic in forensics and security, with corresponding pixel and frequency analysis solutions to detect visual artifacts~\cite{WARIF2016259}. These methods, in addition to early deep generative models, were able to create some inauthentic content. However, results were easily classified as fake or real by humans, as opposed to deep fakes.

\subsection{GAN Empowerment}

Following GANs proposed by Goodfellow et al. \cite{GAN}, deep learning models have been advancing in generative tasks for inpainting \cite{Iizuka17}, translation \cite{cyclegan}, and editing \cite{neuralediting}.
Inherently, all generative approaches suffer from the control over generation. In the context of GANs, this problem is mostly explored by Variational Autoencoders (VAE) \cite{kingma16} and Conditional GANs \cite{mofa} to control the generation by putting constraints in the latent space. In addition to the improvements in controlling GANs, other approaches improved training efficiency, accuracy, and realism of GANs by deep convolutions \cite{dcgan}, Wasserstein distances \cite{wass}, least squares \cite{lsgan}, and progressive growing \cite{pggan}. It is arguable that these developments, in addition to the availability of such models, seeded the authenticity problem.

\begin{figure*}[hb]
\centering
  \includegraphics[width=1\linewidth]{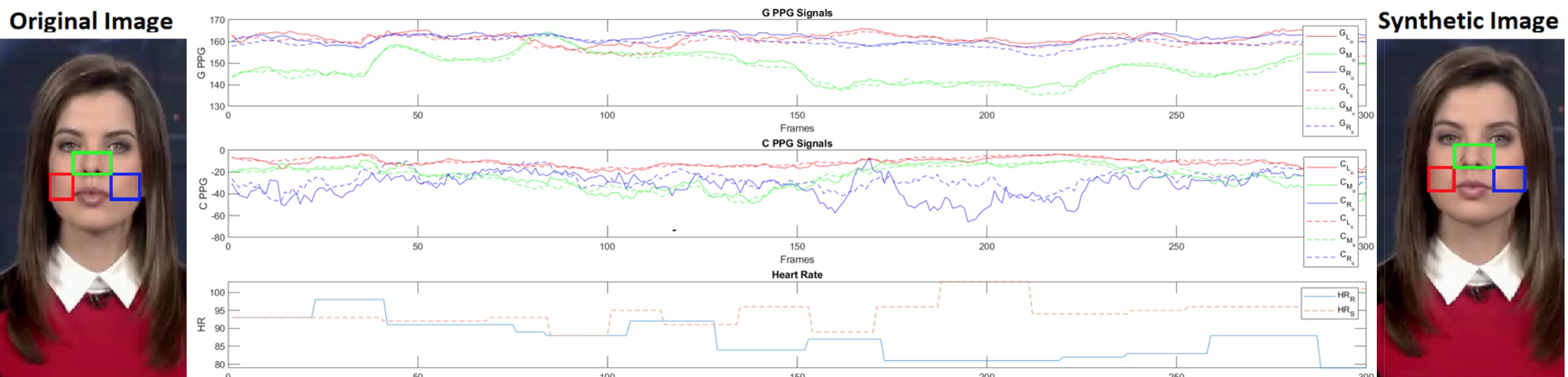}
 \caption{\textbf{Biological Signals.} Green ($G_*$ - top) and chrom-PPG ($C_*$ - middle) from left ($*_L$ - red), middle ($*_M$ - green), and right ($*_R$ - blue) regions.  Heart rates ($HR_*$ - bottom), and sample frames of original (left, $*_{*_O}$ - solid) and synthetic (right, $*_{*_S}$ - dashed) videos.} 
  \label{fig:signals}
\end{figure*}
\subsection{Synthetic Faces}
Since Viola-Jones \cite{VJ}, computer vision community treasures the domain of facial images and videos as one of the primary application areas. Therefore, numerous applications and explorations of GANs emerged for face completion \cite{gfc}, facial attribute manipulation \cite{shen17,stargan18,attgan17}, frontal view synthesis \cite{huang17}, facial reenactment \cite{deepvideoportrait, f2f, xu17}, identity-preserving synthesis \cite{bao18}, and expression editing \cite{exprgan}. In particular, advancements in generative power, realism, and efficiency of VAEs and GANs for facial reenactment and video synthesis resulted in the emergence of the ``deep fake'' concept, which is replacing the face of a target person with another face in a given video, as seamless as possible. The exact approach is not published, however the deep fake generator is assumed to consist of two autoencoders trained on source and target videos: Keeping the encoder weights similar, so that general features can be embedded in the encoder and face-specific features can be integrated by the decoder. Another approach, Face2Face \cite{f2f}, reconstructs a target face from a video and then warps it with the blend shapes obtained by the source video in real-time. Deep Video Portraits \cite{deepvideoportrait} and vid2vid~\cite{vid2vid} follow this approach and employ GANs instead of blend shapes. Overall, the results are realistic, but there are still skipped frames and face misalignments due to illumination, occlusion, compression, and sudden motions.

\subsection{Image Forensics}
In par with the increasing number of inauthentic facial images and videos, methods for detecting such have also been proposed. Those are mostly based on finding inconsistencies in images, such as detecting distortions \cite{boulkenafet16}, finding compression artifacts \cite{barni17}, and assessing image quality \cite{galbally14}. However, for synthetic images in our context, the noise and distortions are harder to detect due to the non-linearity and complexity of the learning process~\cite{facefeatures}. 

There exist two different strategies to tackle this problem, (i) pure deep learning based approaches that act as a detector of a specific generative model~\cite{mesonet, shallownet, 8639163, 8553251, 8124497, 8014963, 8682602}, and (ii) semantic approaches that evaluate the generated faces' realism~\cite{8638330, 8553270, blink}. We summarize all of these in Table~\ref{tab:forensics}. The methods in the first branch investigate the color and noise distributions of specific networks \cite{colordisparity, dctforgery, Li_2019_CVPR_Workshops} or specific people~\cite{Agarwal_2019_CVPR_Workshops}, or train CNNs blindly for synthetic images \cite{mesonet, shallownet, 8682602}. However, they are unfit to be accepted as general synthetic portrait video detection mechanisms, as they rely heavily on detecting artifacts inherent to specific models. Semantic approaches, on the other hand, utilize inconsistencies in the biological domain, such as facial attributes \cite{8638330, 8683164}, mouth movement inconsistency \cite{8553270}, and blink detection \cite{blink}. Our motivation follows the second stream, however we explore real signals instead of physical attributes. Thus our input is continuous, complex, and stable; making our system embrace both perspectives.

\begin{table}[h]
  \centering
\begin{tabular}{c|c|c|c}
Ref. & Accuracy & Dataset & Limitation
\\\hline
\cite{mesonet}   & 98\%         & Own, NA & only \cite{DeepFakes} \\
\cite{mesonet}   & 95\%         & FF\cite{ff} & only \cite{f2f}  \\
\cite{shallownet}& 93.99 AUROC  & celebA\cite{celeba} \& \cite{pagan}& image only\\
\cite{8639163}   & 97.1\%       & Own, NA & unpaired\\
\cite{8553251}   & 99.60\%      & FF\cite{ff} & only \cite{f2f} \& image only \\
\cite{8553251}   & 46.39\% ERR  & Own, NA   & only \cite{FaceSwap} \& \cite{SwapMe}, image only \\
\cite{8124497}   & 92\%         & Own, NA   & image only\\
\cite{8014963}   & 0.927 AUC    & Own, NA   & only \cite{FaceSwap} \& \cite{SwapMe}, image only   \\
\cite{8638330}   & 0.851 AUC    & Own, NA   & image only \\
\cite{8638330}   & 0.866 AUC    & FF\cite{ff} & image only \\
\cite{8553270}   & 24.74\% EER  & temp. vidTimit\cite{vidTimid}& audio only\\
\cite{8553270}   & 33.86\% EER  & temp. AMI \cite{8099850}& audio only\\
\cite{8553270}   & 14.12\% EER  & temp. grid \cite{grid}& audio only\\
\cite{blink}     & 99\%         & Own, NA & only 50 videos 
\\
\cite{8682602}   & 99\%  & FF\cite{ff} & only \cite{f2f} \& image only  \\
\cite{Li_2019_CVPR_Workshops}  & 97\%  & UADFV \cite{8683164} & only \cite{8683164} \& image only\\
\cite{Li_2019_CVPR_Workshops}  & 99.9\%  & DF-TIMIT~\cite{DBLP:journals/corr/abs-1812-08685} & only \cite{DBLP:journals/corr/abs-1812-08685} \& image only\\
\cite{Agarwal_2019_CVPR_Workshops}& ~0.95 AUC & Own, NA & person dependent \\
\cite{8683164}   & 0.974 AUROC  & UADFV \cite{8683164} & only \cite{8683164} \& image only \\
\end{tabular}
  \caption{\textbf{Image Forensics.} Recent approaches with their reported accuracies on the corresponding datasets.}
  \label{tab:forensics}
\end{table} 

\subsection{Biological Signals}\label{sec:BioSignals}
The remote extraction of biological signals roots back to the medical community to explore less invasive methods for patient monitoring. Observing subtle changes of color and motion in RGB videos \cite{mitsig,deep} enable methods such as color based remote photoplethysmography (rPPG or iPPG) \cite{ppg,poh10} and head motion based ballistocardiogram (BCD) \cite{bcd}. We mostly focus on photoplethysmography (PPG) as it is more robust against dynamically changing scenes and actors, while BCD can not be extracted if the actor is not still (i.e., sleeping). Several approaches proposed improvements over the quality of the extracted PPG signal and towards the robustness of the extraction process. The variations in proposed improvements include using chrominance features \cite{chromppg}, green channel components \cite{gppg}, optical properties \cite{corppg}, kalman filters \cite{kalman}, and different facial areas \cite{gppg, tulyakov16, corppg, ppg}.

We believe that all of these PPG variations contain valuable information in the context of fake videos. In addition, inter-consistency of PPG signals from various locations on a face is higher in real videos than those in synthetic ones. Multiple signals also help us regularize environmental effects (illumination, occlusion, motion, etc.) for robustness. Thus, we use a combination of G channel-based PPG (or G-PPG, or $G_*$) \cite{gppg} where the PPG signal is extracted only from the green color channel of an RGB image (which is robust against compression artifacts); and chrominance-based PPG (or C-PPG, or $C_*$) \cite{chromppg} which is robust against illumination artifacts.

\subsection{Deep Fake Collections}\label{sec:Datasets}
As the data dependency of GANs increases, their generative power increases, so detection methods' need for generated samples increases. Such datasets are also essential for the evaluation of novel methodologies, such as ours. Based on the release date and the generative methods, there are two generations of datasets which significantly differ in size and source.

The first generation includes (i) UADFV~\cite{8683164}, which contains 49 real and 49 fake videos generated using FakeApp~\cite{FakeApp}; (ii)  DF-TIMIT~\cite{DBLP:journals/corr/abs-1812-08685}, which includes 620 deep fake videos from 32 subjects using facewap-GAN~\cite{FaceSwap-Gan}; and (iii) Face Forensics\cite{ff} (FF) which has original and synthetic video pairs with the same content and actor, and another compilation of original and synthetic videos, all of which are created using the same generative model~\cite{f2f}. The train/validation/test subsets of FF are given as 704 (70\%), 150 (15\%) and 150 (15\%) real/fake video pairs respectively.  

The second generation aims to increase the generative model diversity while also increasing the dataset size. Face Forensics++~\cite{FF++} dataset employs 1000 real youtube videos and generates the same number of synthetic videos with four generative models, namely Face2Face~\cite{f2f}, Deepfakes~\cite{DeepFakes}, FaceSwap~\cite{FaceSwap} and Neural Textures~\cite{10.1145/3306346.3323035}. Google/Jigsaw dataset~\cite{GoogleDataset} contains 3068 synthetic videos based on 363 originals of 28 individuals using an unrevealed method. Lastly, Celeb-DF~\cite{Celeb_DF_cvpr20} dataset consists of 590 real and 5639 synthetic videos of 59 celebrities, generated by swapping faces using \cite{DeepFakes}. We point out the class imbalance for this dataset, as we will explore its side effects in some results. 

For our experiments, analysis, and evaluation, we use two datasets from each generation; namely UADFV, FF, FF++, and Celeb-DF, in addition to our own Deep Fakes Dataset.

\section{Biological Signal Analysis on \\Fake \& Authentic Video Pairs}

We employ six signals $S = \{G_L, G_R, G_M, C_L, C_R, C_M\}$ that are combinations of G-PPG \cite{gppg} and C-PPG \cite{chromppg} on the left cheek, right cheek \cite{corppg}, and mid-region \cite{tulyakov16}. Each signal is named with channel and face region in subscript. Figure~\ref{fig:signals} demonstrates those signals extracted from a real-fake video pair, where each signal is color-coded with the facial region it is extracted from. We declare signals in Table~\ref{tab:signals} and transformations in Table~\ref{tab:transformations}.

\begin{table}[h]
  \centering
\begin{tabular}{c|c}
Symbol & Signal
\\\hline
$S$ & $\{G_L,G_R,G_M,C_L,C_R,C_M\}$ \\
$S_C$ & $\{C_L,C_R,C_M\}$ \\
$D$ & $\{|C_L-G_L|,|C_R-G_R|,|C_M-G_M|\}$ \\
$D_C$ & $\{|C_L-C_M|, |C_L-C_R|, |C_R-C_M|\}$ \\
\end{tabular}
  \caption{\textbf{Signal Definitions.} Signals used in all analysis.}
  \label{tab:signals}
\end{table}

\begin{figure*}[hb!]
\centering
  \includegraphics[width=1\linewidth]{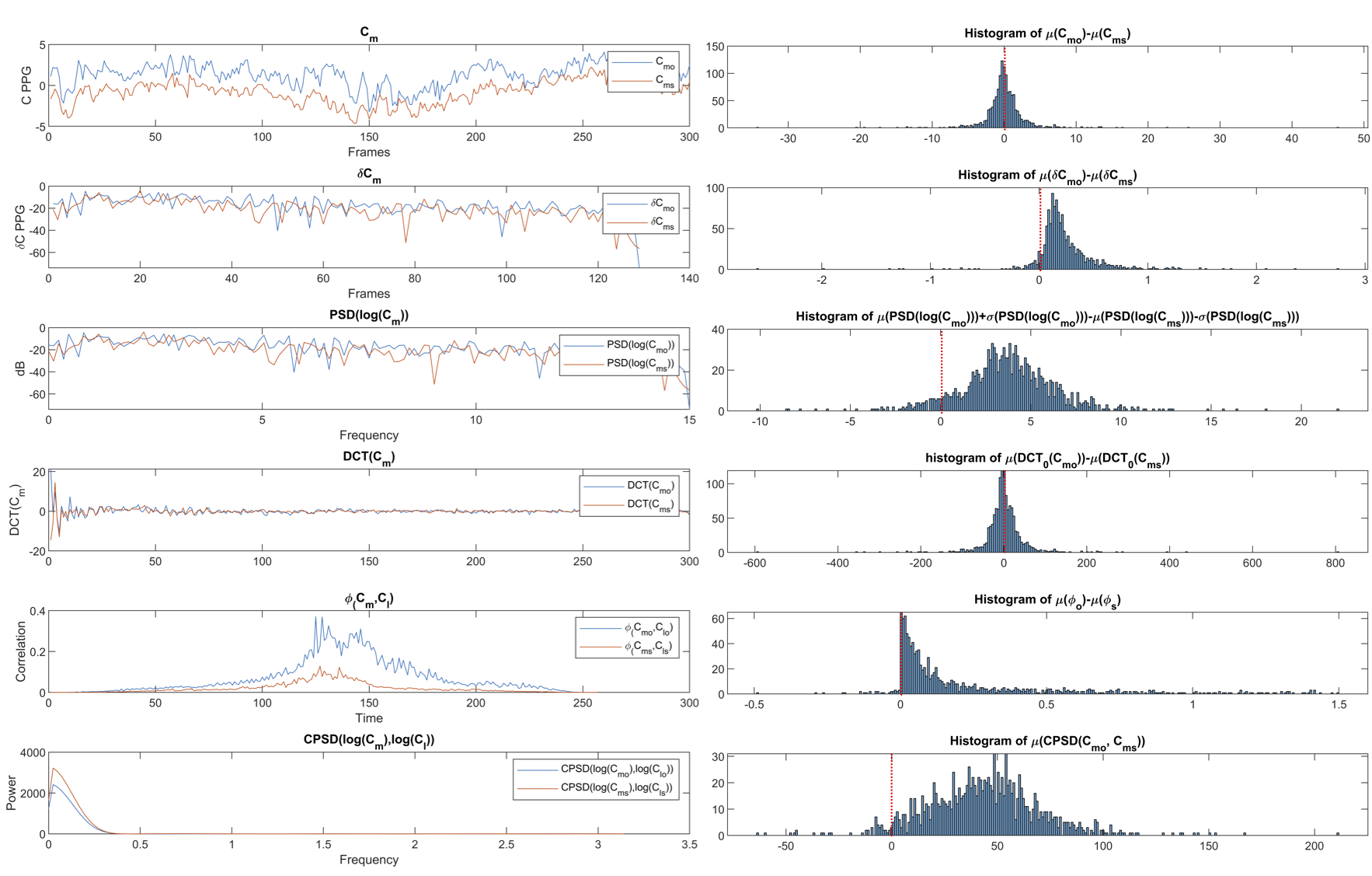}
 \caption{\textbf{Pairwise Analysis.} Example original and synthetic signal pair $C_{M_o}$ and $C_{M_s}$, their derivatives $\delta C_M$, power spectral densities $P(L(C_M))$, discrete cosine transforms $\mathrm{X}(C_M)$, cross correlation $\phi(C_M, C_L))$, and cross power spectral density $A_p(S_C)$ (left). Histograms of mean differences of these values for all pairs in the dataset (right).}
  \label{fig:pairwiseImage2}
\end{figure*}

In order to understand the nature of biological signals in the context of synthetic content, we first compare signal responses in original ($PPG_o$) and synthetic ($PPG_s$) video pairs using traditional signal processing methods: log scale ($L$), Butterworth filter \cite{butter} ($H$), power spectral density ($P$), and combinations (Figure~\ref{fig:pairwiseImage}). We start by comparing signals (top) and their derivatives (bottom) and formulate that distinction (i.e., fourth column shows contrasting structure for their power spectra). Here, the preservation of spatial coherence and temporal consistency is observed in authentic videos, and an error metric that encapsulates these findings in a generalized classifier is desired. This analysis also sets ground for understanding generative systems in terms of biological replicability. 
\begin{table}[h]
  \centering
\begin{tabular}{c|c}
Symbol & Transformation 
\\\hline
$A(S)$ & autocorrelation \\
$\hat{A}(S)$ & spectral autocorrelation \\
$\phi(S_x, S_y)$ &cross correlation \\
$P(S)$ & power spectral density \\ 
$A_p(S_C)$ & pairwise cross 
spectral densities \\
$L(S)$ & log scale \\
$\mathrm{X}(S)$ & discrete cosine transform \\
$W(S)$ & Wavelet transform \\
$Y(S)$ & Lyapunov function~\cite{wc} \\
$G(S)$ & Gabor-Wigner transform~\cite{gwt} \\
\end{tabular}
  \caption{\textbf{Transformation Definitions.} Transformation functions used throughout the analysis of signals specified in Table~\ref{tab:signals}.}
  \label{tab:transformations}
\end{table}

\subsection{Statistical Features}\label{sec:main}
\begin{figure*}[h]
\centering
  \includegraphics[width=1\linewidth]{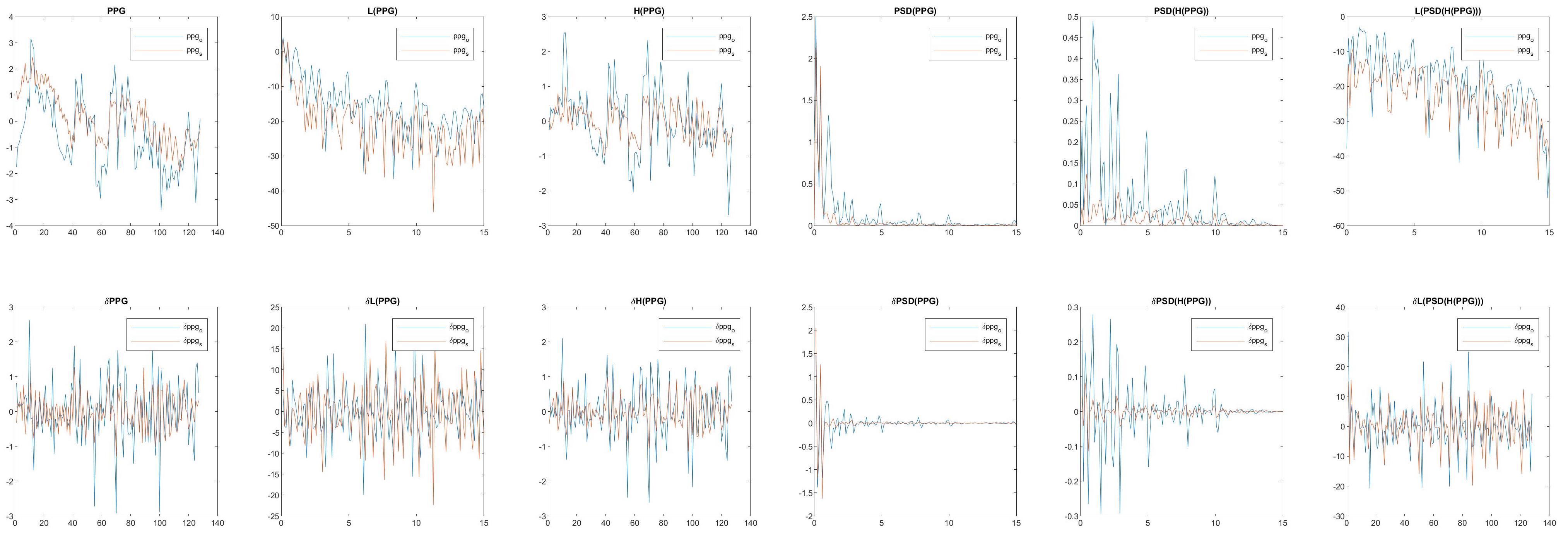}
 \caption{\textbf{Raw Signals.} Characteristics of biological signals (top), same analysis on derivative of signals (bottom), from original (blue) and synthetic (orange) video pairs.}
  \label{fig:pairwiseImage}
\end{figure*}

We set our first task as the pairwise separation problem: Given pairs of fake and authentic videos without their labels, can we take advantage of biological signals for labeling these videos? We use 150 pairs in the provided test subset of Face Forensics \cite{ff} as a base, splitting each video into $\omega$-length temporal segments (the value for $\omega$ is extensively analyzed in Section \ref{sec:segment}). Our analysis starts by comparing simple statistical properties such as mean($\mu$), standard deviation($\sigma$), and min-max ranges of $G_M$ and $C_M$ from original and synthetic video pairs. We observed the values of simple statistical properties between fake and real videos and selected the optimal threshold as the valley in the histogram of these values (Figure~\ref{fig:pairwiseImage2}, first row). By simply thresholding, we observe an initial accuracy of 65\% for this pairwise separation task. Then, influenced by the signal behavior (Figure~\ref{fig:pairwiseImage}), we make another histogram of these metrics on all absolute values of differences between consecutive frames for each segment (i.e., $\mu(|G_M(t_n)-G_M(t_{n+1})|)$), achieving 75.69\% accuracy again by finding a cut in the histogram (Figure~\ref{fig:pairwiseImage2}, second row). Although histograms of our implicit formulation per temporal segment is informative, a generalized detector can benefit from multiple signals, multiple facial areas, multiple frames in a more complex space. Instead of reducing all of this information to a single number, we conclude that exploring the feature space of these signals can yield a more comprehensive descriptor for authenticity.

\begin{table*}[hb!]

 \centering
\begin{tabular}{c|c|c}
 Feature & Explanation & Reference \\\hline
 $F_1$ & mean and maximum of cross spectral density & Sec.~\ref{sec:pairwise}\\\hline 
 
 $F_2$ & RMS of differences, std., mean of absolute differences, ratio of negative differences, & \cite{soleymani12}\\ &  zero crossing rate, avg. prominence of peaks, std. prominence of peaks, avg. peak width,  &\\ & std. peak width, max./min. derivative, mean derivative, mean spectral centroid & \\\hline 
 
 $F_3$ & nb of narrow pulses in spectral autocorrelation, nb of spectral lines in spectral & \cite{CR} \\ &  autocorrelation, average energy of narrow pulses, max. spectral autocorrelation & \\ \hline
 
 $F_4$ & std., std. of mean values of 1 sec windows, RMS of 1 sec differences, mean std. of differences& \cite{hrv} \\ &  std. of differences, mean of autocorrelation, Shannon entropy & \\\hline
 
 $F_5$ & first n Wavelet coefficients $W(S)$& \cite{wpc,wc} \\\hline
 
 $F_6$ & largest n Lyapunov exponents $Y(S)$ & \cite{wc} \\\hline
 
 $F_7$ & max. of spectral power density of normalized centered instantaneous amplitude, std. of & \cite{MC}\\ & abs. value of the centered non-linear component of instantaneous phase, std. of centered &\\ & non-linear component of direct instantaneous phase, std. of abs. value of normalized centered  & \\ & instantaneous amplitude, kurtosis of the normalized instantaneous amplitude & \\\hline 
 
 $F_8$ & log scale power of delta (1-4HZ), theta (4-8HZ), and alpha (8-13HZ) bands & \cite{ER}\\\hline
 
 $F_9$ & mean amplitude of high frequency signals, slope of PSD curves between high and low & \cite{gitthesis} \\ & frequencies, variance of inter-peak distances &\\\hline

\end{tabular}
\caption{\textbf{Feature Sets.} Feature sets (left) from all experiments are explained (middle) and documented by a reference (right).}\label{tab:features}
\end{table*}
\subsection{Power Spectra}
In addition to analyzing signals in time domain, we also investigate their behavior in frequency domain. Thresholding their power spectrum density ($P(S)$) in linear and log scales results in an accuracy of 79.33\% (similar to Figure~\ref{fig:pairwiseImage2}, third row) using the following formula:
\begin{equation}
    \mu_{P(G_{L_o})}+\sigma_{P(G_{L_o})}-(\mu_{P(G_{L_s})}+\sigma_{P(G_{L_s}))}
\end{equation}where the definition of $P$ can be found in Table \ref{tab:transformations} and $G_L$ can be found in Table \ref{tab:signals}, and subscripts denote $G_{L_o}$ for original and $G_{L_s}$ for synthetic. We also analyze discrete cosine transforms (DCT) ($\mathrm{X}$) of the log of these signals. Including DC and first three AC components, we obtain 77.41\% accuracy (Section \ref{SEC_DCT}). We further improve the accuracy to 91.33\% by using only zero-frequency (DC value) of $\mathrm{X}$. 

\subsection{Spatio-temporal Analysis}\label{sec:pairwise}
Combining previous two sections, we also run some analysis for the coherence of biological signals within each signal segment. For robustness against illumination, we alternate between $C_L$ and $C_M$ (Table \ref{tab:signals}), and compute cross-correlation of their power spectral density as $\phi(P(C_M),P(C_L))$. Comparing their maximum values gives 94.57\% and mean values gives 97.28\% accuracy for pairwise separation. We improve this result by first computing power spectral densities in log scale $\phi(L(P(C_M)),L(P(C_L)))$ (98.79\%), and even further by computing cross power spectral densities $\mu(A_p(L(S_{C_o}))-A_p(L(S_{C_s})))$ (99.39\%). Last row in Figure~\ref{fig:pairwiseImage2} demonstrates that difference, where 99.39\% of the pairs have an authentic video with more spatio-temporally coherent biological signals. This final formulation results in an accuracy of 95.06\% on the entire Face Forensic dataset (train, test, and validation sets), and 83.55\% on our Deep Fakes Dataset. 

\section{Generalized Content Classifier}
We hypothesize that our metric to separate pairs of original and synthetic videos with an accuracy of 99.39\% is a promising candidate to formulate the inconsistency into a generalized binary classifier. In the pairwise setting, comparison of aggregate spatio-temporal features are representative enough. However, as these signals are continuous and noisy, there is no universal hard limit to robustly classify such content. To build a generalized classifier, we experiment with several signal transformations in time and frequency domains (as defined in Table~\ref{tab:transformations}) to explore the artifacts of synthetic content towards characteristic feature sets (Table~\ref{tab:features}).

\subsection{Feature Sets}
We explored several features to be extracted from the signals declared in Table \ref{tab:transformations}. Due to the fact that rPPG is mostly evaluated by the accuracy in heart rate, we consult other features used in image authenticity \cite{facefeatures}, classification of Electroencephalography (EEG) signals \cite{wpc, ER}, statistical analysis \cite{MC, soleymani12, CR}, and emotion recognition \cite{wc, ER}. These feature sets are enumerated in Table \ref{tab:features} together with the reference papers for biological signal classification. We exhaustively document all possible feature extractors in the literature for robustness and we refer the reader to specific papers for the formulation and explanation of the features.

\subsection{Authenticity Classification}\label{sec:auth}
Following our motivation to understand biological signals in the context of fake content, we covet interpretable features with no assumption on the fake content. This is the key that leads us to employ support vector mahine (SVM) with a radial basis function (RBF) kernel~\cite{svm} for this binary classification task, instead of a DNN. We conduct many experiments by training an SVM using feature vectors extracted from the training set, and then report the accuracy of that SVM on the test set. All of these experiments are denoted with $F_*(T(S))$ where $F_*$ is the feature extractor from Table~\ref{tab:features} applied to (transformed) signal $T(S)$ from Table~\ref{tab:transformations}. Both signal transformation and feature extraction can be applied to all elements of the inner set.

For exploration, we combine all subsets of Face Forensics (FF) dataset \cite{ff} and randomly split the combined set to train (1540 samples, 60\%) and test sets (1054 samples, 40\%). We create feature vectors with maximum and mean ($F_1$) of cross power spectral densities of $S_C$ ($A_p(S_C)$) for all videos in the train set, as it was the feature with the highest accuracy from Section~\ref{sec:pairwise}. Unlike pairwise results, SVM accuracy with $f=F_1(A_p(S_C))$ is low (68.93\%) but this sets a baseline for next steps. Next, we classify by $f=\mu_{P(S)}$ (six features per sample) achieving 68.88\% accuracy, and by $f=\mu_{A_p(D_C)}\cup \mu_{A(S)}$ (9 features) achieving 69.63\% accuracy on the entire FF dataset. Table~\ref{tab:exp} lists 7 of 70 experiments done using a combination of features. These chronologically higher results indicate the driving points of our experimentation, each of these led us to either pursue a feature set, or leave it out completely. The other 63 experiments, both on FF dataset and our Deep Fakes Dataset, are listed in Section \ref{SEC_APPA}.

\begin{table}[hb]
    \centering
    \begin{tabular}{c|c|c}
        $f$ & $|f|$ & $\tilde{f}$ \\\hline
        $F_3(\hat{A}(S))$ & $4\times6$ &  67.55\% \\\hline
        $F_6(L(S))$ & 600 &  69.04\% \\\hline
        $F_4(log(S))$ & 60 & 69.07\% \\\hline
        $F_2(S)$ & $13\times6$ & 69.26\% \\\hline
        $F_5(P(W(S)))$ & 390 & 69.63\% \\\hline
        $F_4(S)\cup$ &$6\times6\ +$& \\
        $F_3(log(S))\cup$ & $4\times6\ +$ & 71.34\% \\
        $\mu_{A_p(D_C)}$ & 3& \\\hline
        $F_4(log(S)\cup A_p(D_C))\cup $ &$6\times9\ +$& \\ 
        $\cup\ F_1(log(D_C))\cup$ & $6\ +$ & 72.01\% \\ 
        $F_3(log(S)\cup A_p(D_C))$ &$4\times9$& \\\hline 
    \end{tabular}
    \caption{\textbf{Experiments.} $<$ $F_n$ from Table \ref{tab:features}$>$ ($<$transformation from Table \ref{tab:transformations}$>$ ($<$signal from Table \ref{tab:signals}$>$)) (left), size of feature vector (middle), and highest achieved segment classification accuracy (right) of some experiments.}
    \label{tab:exp}
\end{table}

Based on our experiments, we conclude that ``authenticity'' (i) is observed both in time and frequency domains, (ii) is highly sensitive to small changes in motion, illumination, and compression if a single signal source is used, and (iii) can be discovered from coherence and consistency of multiple biological signals. We infer conclusion (i) from high classification results when $A(.)$, $\hat{A}(.)$, and $F_4$ in time domain is used in conjunction with $P(.)$, $A_p(.)$, and $F_3$ in frequency domain; (ii) from low true negative numbers when only $S$ is used instead of $D_C$ or $D$, or only using $F_5$ or $F_6$, and (iii) from high accuracies when $D_C$, $\phi(.)$, $A_p(.)$, and $F_1$ is used to correlate multiple signals  (Table~\ref{tab:exp}). Our system is expected to be independent of any generative model, compression/transmission artifact, and content-related influence: \textit{a robust and generalized FakeCatcher, based on the essence of temporal consistency and spatial coherence of biological signals in authentic videos}. Our experimental results conclude on the following implicit formulation,

\begin{equation*}
    \begin{aligned}
    f =\ & F_1(log(D_C))\ \cup \\
     \ & F_3(log(S)\ \cup A_p(D_C))\  \cup \\
     \ & F_4(log(S)\ \cup A_p(D_C))\ \cup \\
     \ & \mu\hat{A}(S) \  \cup max(\hat{A}(S))
\end{aligned}
\end{equation*}
where we gathered 126 features combining $F_1, F_3, F_4$ sets, on log transforms, pairwise cross spectral densities, and spectral autocorrelations, of single source $S$ and multi-source $D_C$ signals. The SVM classifier trained with these 126 features on the FF dataset results in 75\% accuracy. We also perform the same experiment on our Deep Fakes Dataset (with a 60/40 split) obtaining an accuracy of 76.78\%.

\subsection{Probabilistic Video Classification}
As we coulAs mentioned in Section~\ref{sec:main}, the videos are split into $\omega$-interval segments for authenticity classification. Considering that our end goal is to classify videos, we aggregate the segment labels into video labels by majority voting. Majority voting increases the segment classification accuracy of 75\% to 78.18\% video classification accuracy within Face Forensics dataset, hinting that some hard failure segments can be neglected due to significant motion or illumination changes.
Consequently, a weighted voting scheme compensates the effect of these erroneous frames. In order to achieve that, we need the class confidences instead of discrete class labels, thus we convert our SVM to a support vector regression (SVR). Instead of a class prediction of SVM, SVR allowed us to learn the probability of a given segment to be real or synthetic. We use the expectation of the probabilities as the true threshold to classify authenticity. Also, if there is a tie in majority voting, we weigh it towards the the expectation. Using the probabilistic video classification, we increase the video classification accuracy to 82.55\% in Face Forensics and to 80.35\% in Deep Fakes Dataset. 

\subsection{CNN-based Classification}
Investigating the failure cases of our probabilistic authenticity classifier, we realize that our misclassification rate of marking real segments as synthetic segments (false positive, FP) is higher than our misclassification rate of marking synthetic segments as real segments (false negative, FN). The samples ending up as false positives contain artifacts that corrupt PPG signals, such as camera movement, motion and uneven illumination. In order to improve the resiliency against such artifacts, we postulate that we can exploit the coherence of signals by increasing the number of regions of interests (ROIs) within a face. We hypothesize that coherence will be more strongly observed in real videos, as artifacts tend to be localized into specific regions in the face. However packing more signals will exponentially grow the already complex feature space (Section~\ref{sec:blind}) for our authenticity classifier. Therefore, we switch to a CNN-based classifier, which is more suitable for a higher-dimensional segment classification task (Figure~\ref{fig:DeepModel}).

\subsubsection{PPG Maps}
Similar to Section~\ref{sec:auth}, we extract $C_M$ signals from the mid-region of faces, as it is robust against non-planar rotations. To generate same size subregions, we map the non-rectangular region of interest (ROI) into a rectangular one using Delaunay Triangulation~\cite{Fortune:1997:VDD:285869.285891}, therefore each pixel in the actual ROI (each data point for $C_M$) corresponds to the same pixel in the generated rectangular image. We then divide the rectangular image into 32 same size sub-regions. For each of these sub-regions, we calculate $C_M=\{C_{M_0}, \dots, C_{M_{\omega}}\}$, and normalize them to $[0, 255]$ interval. We combine these values for each sub-region within $\omega$ frame segment into an $\omega\times32$ image, called PPG map, where each row holds one sub-region and each column holds one frame. Example real and synthetic PPG maps are shown in Figure~\ref{fig:ExamplePPG}, in the first two rows.

\subsubsection{Learning Authenticity}\label{sec:cnn}
\begin{figure}[h]
\centering
  \includegraphics[width=1\linewidth]{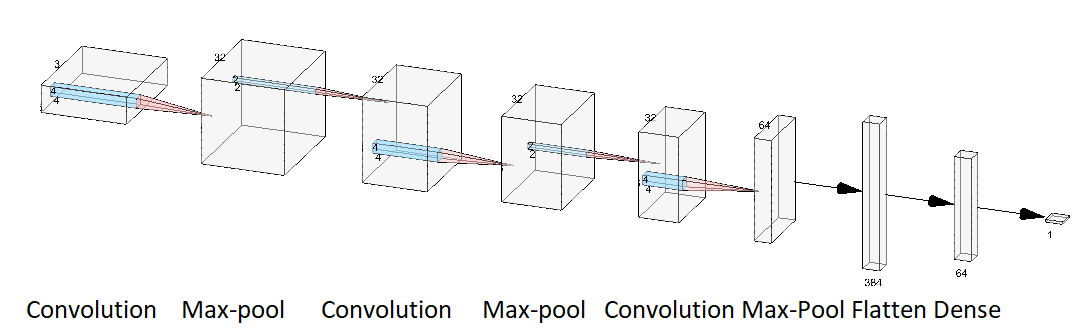}
 \caption{\textbf{CNN Architecture.} Three convolutional layers with max pooling, followed by dense layers.}
  \label{fig:DeepModel}
\end{figure}

We use a simple three layer convolutional network with pooling layers in between and two dense connections at the end (Figure ~\ref{fig:DeepModel}). We use ReLU activations except the last layer, which is a sigmoid to output binary labels. We also add a dropout before the last layer to prevent overfitting. We do not perform any data augmentation and feed PPG maps directly. Our model achieves 88.97\% segment and 90.66\% video classification accuracy when trained on FF train set and tested on the FF test set with $\omega=128$. Similarly, our model obtains 80.41\% segment and 82.69\% video classification accuracy when trained on our Deep Fakes Dataset with a random split of $60/40$.  

\subsubsection{Spectral PPG Maps}
\begin{figure}[h!]
\centering
  \includegraphics[width=1\linewidth]{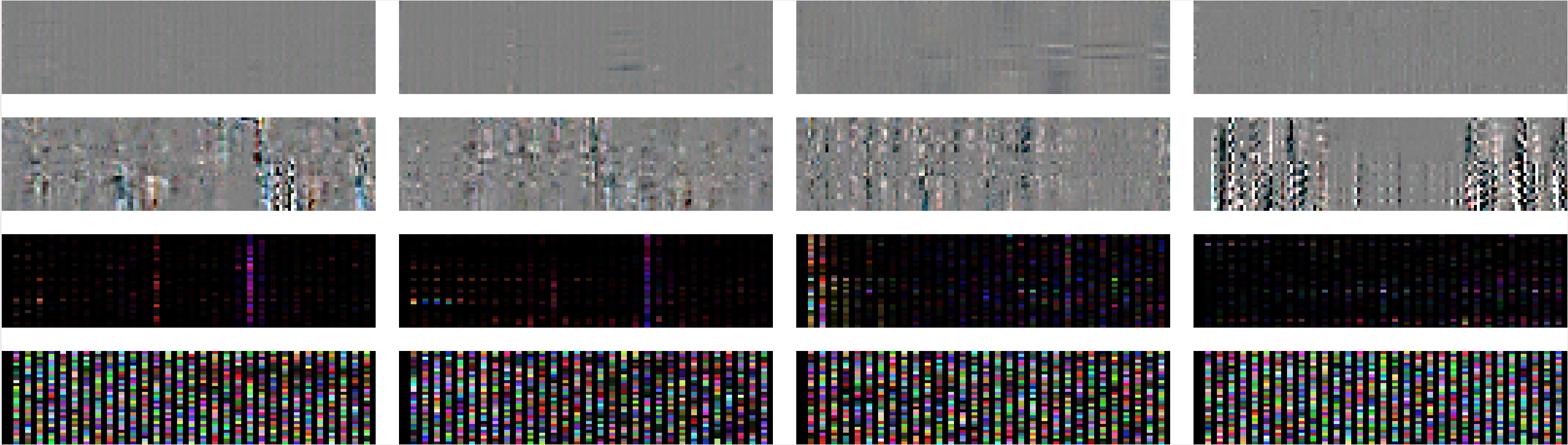}
 \caption{\textbf{PPG Maps}. PPG signals per sub-region per frame are converted into $128\times32$ images for five synthetic (top) and original (bottom) segments.}
  \label{fig:ExamplePPG}
\end{figure}

As it is discussed in Section~\ref{sec:pairwise}, frequency domain also holds important consistency information for detecting authentic content. Thus, we enhance our PPG maps with the addition of encoding binned power spectral densities $P(C_M)=\{P(C_M)_0, \dots, P(C_M)_{\omega}\}$ from each sub-region, creating $\omega\times64$ size images. Examples of our real and synthetic spectral PPG maps are shown in the last two rows of Figure~\ref{fig:ExamplePPG}. This attempt to exploit temporal consistency improves our accuracy for segment and video classification to 94.26\% and 96\% in Face Forensics, and 87.42\% and 91.07\% in Deep Fakes Dataset. Further classification results on different datasets are reported in Tables~\ref{tab:ClassBased} and \ref{tab:CrossDataset}, such as 91.50\% on Celeb-DF, 94.65\% on FF++, and 97.36\% on UADFV. 

\section{Results}
Our system utilizes Matlab~\cite{MATLAB:2010} for signal processing, Open Face library~\cite{openface} for face detection, libSVM~\cite{Chang:2011:LLS:1961189.1961199} for the classification experiments, Wavelab~\cite{wavelab} for Wavelet transformation and $F_5$ feature set, and Keras~\cite{keras} for network implementations. In this section, we first describe the benchmark datasets and introduce our new Deep Fakes Dataset. Then we examine some parameters of the system, such as the facial regions of signals in $S$, segment durations $\omega$, and dimensionality reduction techniques on the feature set $F$. We also compare our system against other detectors, demonstrate that our system is superior to complex baseline architectures, and summarize our experimental outcomes.

\subsection{Deep Fakes Dataset}
The external datasets we used are explained in Section\ref{sec:Datasets}. Although FF is a clean dataset perfect for initial experiments, we need to assess the generalizability of our findings. For this purpose, we created a new database (so-called Deep Fakes Dataset (DF)), which is a set of portrait videos collected ``in the wild''. The videos in our dataset are diverse real-world samples in terms of the source generative model, resolution, compression, illumination, aspect-ratio, frame rate, motion, pose, cosmetics, occlusion, content, and context, as it originates from various sources such as media sources, news articles, and research presentations; totaling up to 142 videos, ~32 minutes, and ~30 GBs. For each synthetic video, we searched for the original counterpart if it is not presented with its source. For example, if a synthetic video is generated from a movie, we found the real movie and cropped the corresponding segment. When we could not locate the source, we included a similar video to preserve the class balance (i.e., for a given fake award speech, we included a similar (in size, location, and length) original award speech). This variation does not only increase the diversity of our synthetic samples, but it also makes our dataset more challenging by breaking possible real-fake associations.

\begin{figure}[h]
\centering
  \includegraphics[width=1\linewidth]{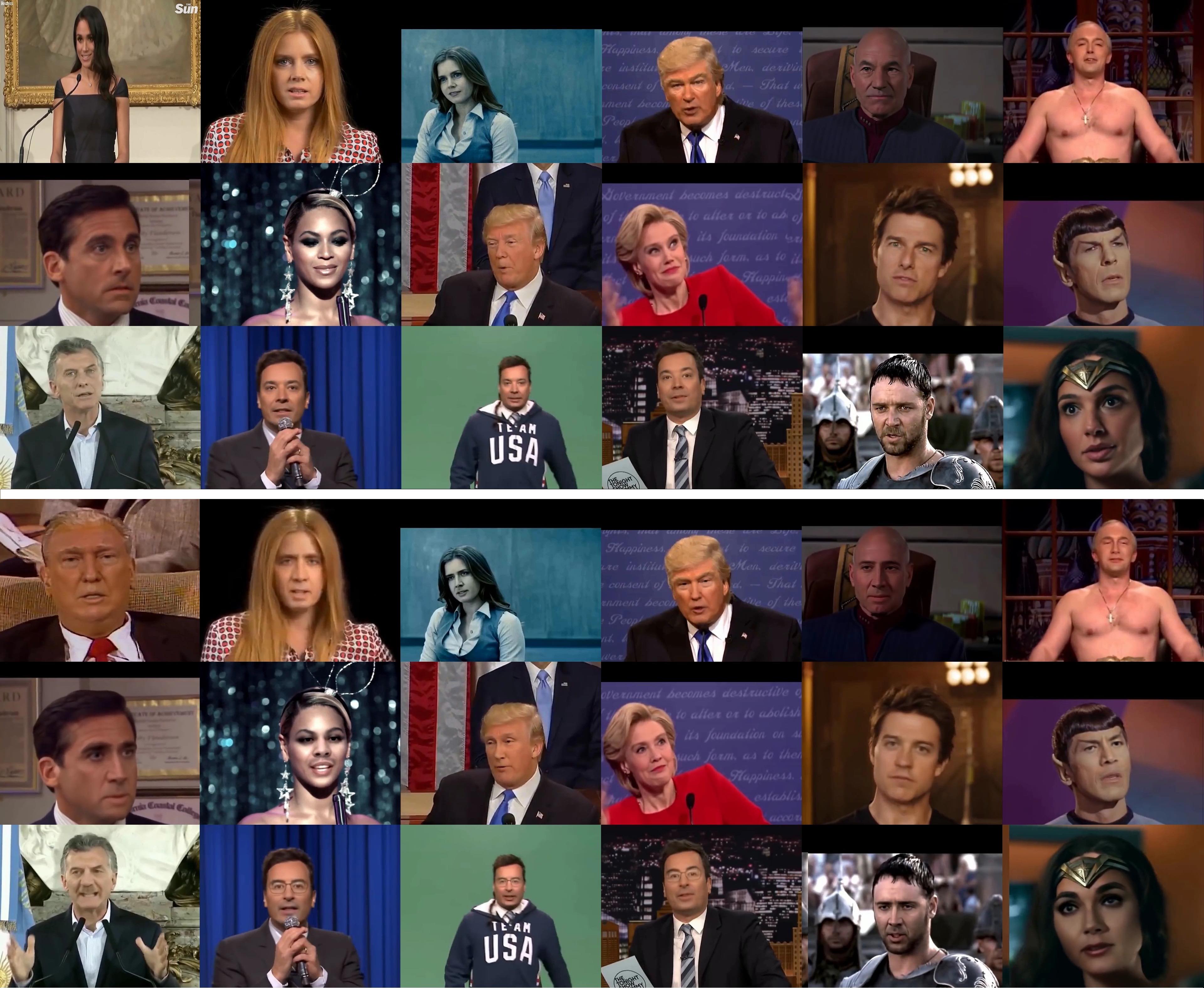}
 \caption{\textbf{Deep Fakes Dataset.} We introduce a diverse dataset of original (top) and fake (bottom) video pairs from online sources.}
  \label{fig:DeepFakesExamples}
\end{figure}

Figure~\ref{fig:DeepFakesExamples} demonstrates a subset of the Deep Fakes Dataset, original videos placed in the top half and fakes in the bottom half. A small clip consisting of several videos in the dataset can also be found in the Supplemental Material. The dataset is publicly released for academic use\footnote{http://bit.ly/FakeCatcher}. High accuracy on Deep Fakes Dataset substantiates that FakeCatcher is robust to low-resolution, compression, motion, illumination, occlusion, pose, and cosmetics artifacts; that enrich the input and slightly reduce the accuracy, without preprocessing. We also perform cross-dataset evaluations which support our aforementioned claims about generalizability.

\subsection{Evaluations}
We conduct several evaluations to establish our approach as state-of-the-art solution for deep fake detection. First, we justify the use of biological signals by several comparisons to off the shelf deep learning solutions and to other fake detectors. Second, we list several crucial experiments to guide us in the combinatorial signal-feature space. Third, we summarize and relate our experimental findings to each result experiment section.

\subsubsection{Quantitative Comparisons}\label{sec:comp}
In order to promote the effect of biological signals on the detection rate, we perform some experiments with several networks: (i) a simple CNN, (ii) Inception V3 \cite{Szegedy_2016_CVPR}, (iii) Xception \cite{Chollet_2017_CVPR}, (iv) ConvLSTM \cite{NIPS2015_5955}, (v-vii) three best networks proposed in \cite{shallownet}, and (viii) our approach.

\begin{table}[h]
\centering
\begin{tabular}{c|c|c|c}
Model & Frame & Face &  Video\\
\hline
Simple CNN & 46.89\% & 54.56\% & 48.88\% \\
InceptionV3 & 73.85\% & 60.96\% & 68.88\%\\
Xception & 78.67\% & 56.11\% & 75.55\%\\
ConvLSTM & 47.65\%& 44.82\% & 48.83\%\\
\cite{shallownet} V1 & \textbf{86.26\%} & - & 82.22\%\\
\cite{shallownet} V3 & 76.97\% & - & 73.33\%\\
\cite{shallownet} ensemble & 83.18\% & - &80.00\%\\
\hline
Ours & - & \textbf{87.62\%} & \textbf{91.07\%} \\
\end{tabular}
\caption{\textbf{Comparison.} Detection accuracies of several networks trained on images, face images, and videos. The next best after our approach is 8.85\% less accurate.}\label{tab:compnn}
\end{table}

All experiments in Table~\ref{tab:compnn} are performed on the same 60\% train and 40\% test split of our Deep Fakes Dataset, with same meta parameters. We choose to compare on Deep Fakes Dataset, because it is more generalizable as discussed in the previous section. For ConvLSTM and our approach, ``Frame'' and ``Face'' indicate segment accuracy, for others they indicate frame accuracy. The last column is video accuracy (Section 4.3). We did not run \cite{shallownet} on face images because their approach utilizes background, and we did not run ours on entire frames because there is no biological signal in the background. We emphasize that \textbf{FakeCatcher outperforms the best baseline architecture by 8.85\%}. 

\subsubsection{Qualitative Comparisons}
Even though ``deep fakes'' is a relatively new problem, there are a few papers in this domain. \cite{ff} employs a generative model for detection, but their model is restricted to their generative method in \cite{f2f}. \cite{mesonet} also claims a high detection rate if the synthetic content is generated by \cite{f2f} or the VAE used in the \textit{FakerApp}. \cite{blink} reports high accuracy, however their approach is dependent on eye detection and parameterization. All of these \cite{shallownet, mesonet, ff} employ neural networks blindly and do not make an effort to understand the generative noise that we experimentally characterized using biological signals (Section~\ref{sec:exp}).

Based on our comprehensive experiments, we observe that biological signals are not well-preserved in deep fakes (Section~\ref{sec:pairwise}), however, \textit{is the contribution of biological signals significant against pure ML approaches}? We claim that PPG maps encode authenticity using their spatial coherence and temporal consistency. To prove this, we train the CNN in Section~\ref{sec:cnn} with (i) input frames (46.89\%), (ii) faces cropped from input frames (54.56\%), and (iii) our PPG Maps (Section 4.4) (87.62\%) as shown in Table~\ref{tab:compnn}. The significant accuracy increase justifies the use of biological signals. To ensure that this jump is not only the result of temporal consistency, we compare it to the classification accuracy of ConvLSTM on entire and face frames (47.65\% and 44.82\%), which are even lower than frame-based methods. Thus, we certify that (1) \textit{approaches incorporating biological signals are quantitatively more descriptive for deep fake detection compared to pure machine learning based approaches}, (2) \textit{both spatial and temporal properties of biological signals are important}, and (3) \textit{these enable our network to perform significantly better than complex and deeper networks}. 

\subsubsection{Validation Experiments}\label{SEC_APPA}
We experimentally validate the combination of signals and feature sets used throughout the paper, namely 6 signals, in 11 transformations, in 126 features; explored in over 70 experiments. We document all experiments with the signal transformation (Table~\ref{tab:transformations}), feature set (Table~\ref{tab:features}), and SVM classification accuracy, trained on FF (Table \ref{tab:all3}) and on DF datasets (Table \ref{tab:alldf}), with $\omega=128$. Furthermore, in Table~\ref{tab:allffmix}, we document experiments where we selected specific subsets of Face Forensics (FFT for train, FFX for test, and FFC for validation sets).

\begin{table}[h]
\centering
\begin{tabular}{c|c|c|c}
Feature set&Signal&\# features&Accuracy
\\\hline
$F_6$             &S                  &600        &43.22\\\hline
$F_{12}$          &S                  &12         &43.41\\\hline
$F_3$             &log(S)             &24         &43.81\\\hline
$F_6$             &log(S)             &600        &43.81\\\hline
$F_{12}$          &S                  &12         &43.61\\\hline
$F_3$             &S                  &24         &45.57\\\hline
$F_4$             &log(S)             &36         &48.13\\\hline
$F_7$             &S                  &30         &48.52\\\hline
$F_4F_3F_1$       &$log(S)$x2,$log(D_C)$, &96     &54.02\\
$F_4F_3$          &$A_p(log(D_C))$ &&\\\hline
$F_5$             &S                  &767        &54.22\\\hline
$\mu()max()$      &$A(log(S))$        &12         &54.22\\\hline
$F_7$             &log(S)             &30         &55.40\\\hline
$F_5$             &S                  &390        &58.84\\\hline
$F_5$             &A(log(S))          &390        &60.90\\
\end{tabular}
  \caption{\textbf{Classification Results on DF.} Accuracies with $\omega=128$, on DF dataset, and train/test split of 60/40.
}
  \label{tab:alldf}
\end{table}

\begin{table}
\centering
\begin{tabular}{c|c|c|c}
Feature Set&Signal&\#Feat.&Acc
\\\hline
$F_5$ 	        &$S$	    &774	&51.23\\\hline
$F_8$	        &$log(S)$	    &18	    &51.81\\\hline
$F_6$	        &$log(S)$	    &600	&51.89\\\hline
$F_7$	        &$S$	    &30	    &58.06\\\hline
$F_7$	        &$S_C$	    &15	    &59.29\\\hline
$\mu(A_p()))$   &$log(D)$	    &3	    &62.3\\\hline
$F_2$	        &$log(S)$	    &78	    &64.04\\\hline
$F_2$	        &$log(S_C)$	    &39	    &64.32\\\hline
$F_3,F_3$	    &$\hat{A}(S)$,13  &36	    &64.89\\\hline
$F_5$	        &$log(S)$	    &768	&65.27\\\hline
$F_1$	        &$C_L-C_R$	    &2	    &65.37\\\hline
$F_1$	        &$C_L-C_M$	    &2	    &65.55\\\hline
$F_7$	        &$log(S)$	    &30	    &65.55\\\hline

$F_1$	        &$C_M-C_R$	    &2	    &65.84\\\hline
$F_5$	        &$S_C$	    &768	&66.03\\\hline
$F_6$	        &$\hat{A}(S)$	    &600	&66.03\\\hline
$F_2$	        &$S_C$	    &39	    &66.88\\\hline
$\mu(A()))$	    &$S_C$		    &3	    &66.88\\\hline
$\mu(A_p()))$	&$log(D_C)$	    &3	    &66.88\\\hline
$F_5$	        &$A_p(log(D_C))$	    &384	&66.92\\\hline
$F_1$	        &$log(D_C)$	    &6	    &67.17\\\hline
$F_6$	        &$S_C$		    &300	&67.26\\\hline
$F_3$	                    &$log(S)$	                    &24	    &67.55\\\hline
$F_3F_3$	                &$A_p(log(D_C))$,$\hat{A}(S)$	&36	    &67.55\\\hline
$F_4F_3\mu()$	            &$A_p(S)$	                    &63	    &68.12\\\hline
$\mu()$	                    &$A(log(S))$	                &6	    &68.88\\\hline
$F_6$	                    &$log(S)$  	                    &600	&69.04\\\hline
$F_4$	                    &$log(S)$	                    &36	    &69.07\\\hline
$F_2$	                    &$S$	                        &78	    &69.26\\\hline
$F_5$	                    &$P(W(S))$	                    &390	&69.63\\\hline
$\mu()$                     &$A_p(log(D_C))$,$A(log(S))$	&9	    &69.63\\\hline
$F_1F_2F_3F_4F_5$	        &$log(D_C)$,$S$,$log(S)$x2,	    &1944	&70.49\\
$F_5F_6F_7F_{12}$	        &$log(S)$x2,S,$log(S)$x2	    &	&\\\hline
$F_4F_3\mu()$               &$log(S)$x2,$A_p(D_C)$&63	    &71.34\\\hline
$F_4F_3F_1$	                &$log(S)$x2,$log(D_C)$,         &96	&72.01\\
$F_4F_3$                    &$A_p(log(D_C))x2$   &	&\\
\end{tabular}
  \caption{\textbf{Classification Results on FF.} Accuracies of signal transformations and corresponding feature sets.
}
\label{tab:all3}
\end{table}

\subsubsection{Summary of Experimental Outcomes and Findings}\label{sec:exp}

\begin{itemize}
\item \textbf{Spatial coherence:} Biological signals are not coherently preserved in different synthetic facial parts.
\item \textbf{Temporal consistency:} Synthetic content does not contain frames with stable PPG. $\hat{A}$ and $P$ of PPGs significantly differ. However, inconsistency is not separable into frequency bands (Section \ref{SEC_APPB_FREQ}).
\item \textbf{Combined artifacts:} Spatial inconsistency is augmented by temporal incoherence (Experiments in $\ref{SEC_APPA}(S_C)$).
\item \textbf{Artifacts as features:} These artifacts can be captured in explainable features by transforming biological signals (Section \ref{SEC_APPA}, and Tables \ref{tab:signals},\ref{tab:transformations}\&\ref{tab:features}). However there exists no clear reduction of these feature sets into lower dimensions (Section~\ref{sec:blind}), thus CNN performs better than SVM.
\item \textbf{Comprehensive analysis:} Finally, our classifier has higher accuracy for detection in the wild, for shorter videos, and for mid-size ROIs (Section \ref{sec:analysis}).
\end{itemize}

\begin{table*}[hb!]
\centering
\begin{tabular}{c|c|c|c|c|c|c|c}
Feature set&signal&test&$\omega$&train&\# feat.&s. acc&v. acc
\\\hline
$F_5$                                        &$G(S)$                          &FFX&300&FFT+FFC&300&50&-\\
$F_6$                                         &$G(S)$                         &FFX&300&FFT+FFC&600&64.75&-\\
$F_5$                                        &$G(log(S))$                   &FFX&300&FFT+FFC&300&70.25&\\
$LDA_3 (F_1 F_3 F_4$                                  &$log(D_C)$,$A_p(log(D_C))$x2,                       &FFX&300&FFT+FFC&3&71.75&-\\
$F_{12} F_3 F_4)$                                  &$S$,$log(S)$x2                       &&&&&&-\\
    $F_1 F_3 F_4 F_{12} F_3 F_4$              &$log(D_C)$,$A_p(log(D_C))$x2,$S$,$log(S)$x2    &FFC&300&FFT+FFX&108&71.79&-\\
$F_1 F_3 F_4\mu()F_3F_4$&$log(D_C)$,$A_p(log(D_C))$x2,$A(S)$,$log(S)$x2    &FFC&128&FFT+FFX&108&72.21&-\\

$PCA_3(F_1F_3F_4$           &$log(D_C)$,$A_p(log(D_C))$x2,    &FFX&300&FFT+FFC&3&71&69.79\\
$F_{12}F_3F_4)$           &$S$,$log(S)$x2    &&&&&&\\
$F_1F_3F_4F_3F_4F_5$                     &$log(D_C)$,$A_p(log(D_C))$x2,$log(S)$x2,$S$    &FFX&300&FFT+FFC&1631&73.25&72.81\\

$F_1F_3F_4F_{12}F_3F_4$                  &$log(D_C)$,$A_p(log(D_C))$x2,$S$,$log(S)$x2    &FFX&300&FFT+FFC&126&75&75.16\\

$F_1F_3F_4F_{12}F_3F_4$                  &$log(D_C)$,$A_p(log(D_C))$x2,$S$,$log(S)$x2    &FFX&300&FFT&108&74.5&75.50\\
$F_1F_3F_4F_{12}F_3F_4$                  &$log(D_C)$,$A_p(log(D_C))$x2,$S$,$log(S)$x2    &FFX&300&FFT+FFC&108&75&78.18\\
$F_1F_3F_4F_{12}F_3F_4$                  &$log(D_C)$,$A_p(log(D_C))$x2,$S$,$log(S)$x2    &FFX&128&FFT&108&76.37&79.53\\
$F_1F_3F_4F_3F_4$                         &$log(D_C)$,$A_p(log(D_C))$x2,$log(S)$x2     &FFX&128&FFT+FFC&108&75.51&79.53\\
$F_1F_3F_4F_{12}F_3F_4$                  &$log(D_C)$,$A_p(log(D_C))$x2,$S$,$log(S)$x2    &FFX&128&FFT+FFC&126&77.12&81.54\\

$F_1F_3F_4F_{12}F_3F_4$                  &$log(D_C)$,$A_p(log(D_C))$x2,$S$,$log(S)$x2    &FFX&128&FFT+FFC&108&\textbf{77.50}&\textbf{82.55}\\

\end{tabular}
  \caption{\textbf{Classification on mixed train/test sets:} We evaluate FakeCatcher on several subsets, with various signals and features.
}
  \label{tab:allffmix}
\end{table*}

\subsection{Cross Dataset/Model Experiments}
We experiment by training and testing on (i) different datasets, and (i) datasets created by different generative models for the evaluation of our approach. Our cross dataset experiments are conducted between Deep Fakes (ours),   Celeb-DF~\cite{Celeb_DF_cvpr20}, FF~\cite{ff}, FF++~\cite{FF++}, and UADFV~\cite{8683164} datasets where we train our proposed approach on one dataset and test on another (Table~\ref{tab:CrossDataset}). Based on rows 5 and 6, the key observation is that FakeCatcher learns better from small and diverse datasets than on large and single source datasets. On one hand, training on DF and testing on FF is 18.73\% higher than the other way around. On the other hand, DF is approximately only 5\% of FF. The main difference between these datasets is that FF has a single generative model with no other artifacts, and DF is a completely in-the-wild dataset. If we compare rows 3 and 6, we also observe that increasing diversity from FF to FF++, increases the accuracy on DF by 16.9\%.

\begin{table}[h]
\centering
\begin{tabular}{c|c|c}
Train & Test & Video Accuracy
\\\hline
Celeb-DF\cite{Celeb_DF_cvpr20}  & FF++\cite{FF++}                & 83.10\%  \\
FF++\cite{FF++}                 & Celeb-DF\cite{Celeb_DF_cvpr20} & 86.48\%  \\
FF++\cite{FF++}                 & Deep Fakes Dataset                     & 84.51\%  \\
Celeb-DF\cite{Celeb_DF_cvpr20}  & Deep Fakes Dataset                    & 82.39\%  \\
Deep Fakes                      & FF\cite{ff}                    & 86.34\%  \\
FF\cite{ff}                     & Deep Fakes Dataset                     & 67.61\%  \\
FF++\cite{FF++}                 & UADFV\cite{8683164}            & 97.92\%  \\
Deep Fakes Dataset                       & FF++\cite{FF++}                & 80.60\%  \\
Deep Fakes Dataset                       & Celeb-DF\cite{Celeb_DF_cvpr20} & 85.13\%  \\
\end{tabular}
  \caption{\textbf{Cross Dataset Results.} Accuracies for FakeCatcher trained on the first column and tested on the second column.
}
  \label{tab:CrossDataset}
\end{table}

As mentioned in Section~\ref{sec:Datasets}, FF++~\cite{FF++} contains four synthetic videos per original video, where each of them is generated using a different generative model. First, we partition original videos of FF++ into train/test sets with 60/40 percentages. We create four copies of these sets, and delete all samples generated by a specific model from each set (column 1, Table~\ref{tab:CrossModel}, where each set contains 600 real and 1800 fake videos from three models for training, and 400 real and 400 fake videos from one model for test. Table \ref{tab:CrossModel} displays the results of our cross model evaluation. We obtain significantly accurate predictions, except NeuralTextures~\cite{10.1145/3306346.3323035}, as it is an inherently different generative model.

\begin{table}[h]
\centering
\begin{tabular}{c|c|c}
Train & Test & Video Accuracy
\\\hline
FF++ - Face2Face & Face2Face\cite{f2f}                               &95.25\%\\
FF++ - FaceSwap & FaceSwap\cite{FaceSwap}                     &96.25\%\\
FF++ - Deepfakes & Deepfakes\cite{DeepFakes}                   &93.75\%\\
FF++ - NeuralTextures & NeuralTextures\cite{10.1145/3306346.3323035}&81.25\%\\
\end{tabular}
  \caption{\textbf{Cross Model Results on FF++.} Accuracies of FakeCatcher on FF++ dataset variations, where the samples generated by the model in the second column is excluded as test set.
}
  \label{tab:CrossModel}
\end{table}

\begin{figure*}[hb!]
\centering
  \includegraphics[width=1\linewidth]{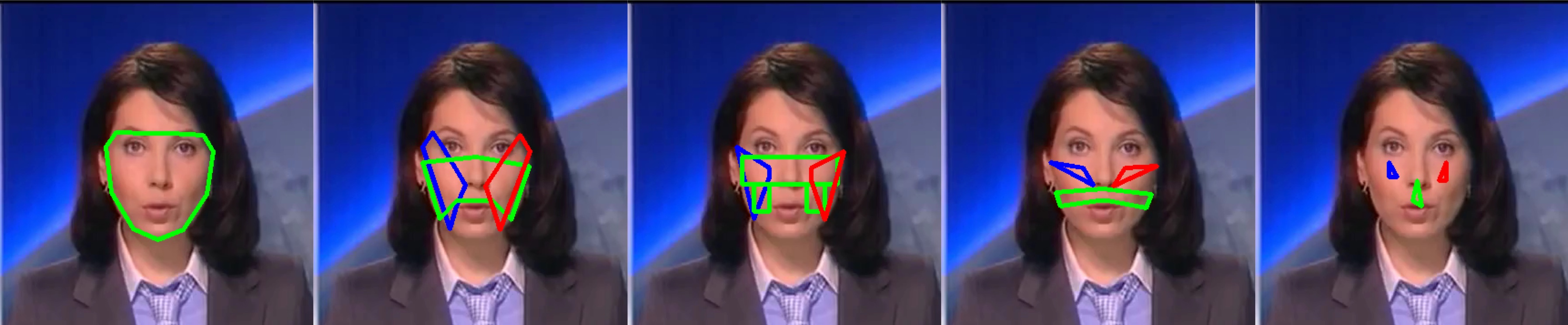}
 \caption{\textbf{ROI Contours.} Whole face (a), big ROI (b), default (c), small ROI (d), few pixels (e), for $G_L$ and $C_L$ (blue), for $G_M$ and $C_M$ (green), and for $G_R$ and $C_R$ (red).}
  \label{fig:roi}
\end{figure*}

\section{Analysis}\label{sec:analysis}
As we introduce some variables into our automatic FakeCatcher, we would like to assess and reveal the best values for those parameters. Segment duration, face region, and different preprocessing techniques for PPGs are explored in our analysis section. Furthermore, we analyzed the explainable feature space for its information content and observed its reaction to dimensionality reduction techniques.

\subsection{Segment Duration Analysis}\label{sec:segment}
Table~\ref{tab:pairw} documents results on the test set, the entire Face Forensics dataset, and Deep Fakes Dataset, using different segment durations. Top half shows the effect of $\omega$ on the pairwise classification. The choice of $\omega=300$ (10 sec) is long enough to detect strong correlations without including too many artifacts for video labels.

Preceded by probabilistic video classification, authenticity classifier can be used with different segment sizes, which we investigate in this section. Selecting a long segment size can accumulate noise in biological signals, in contrast incorrectly labeled segments may be compensated in the later step if we have enough segments. Thus, selecting a relatively smaller segment duration $\omega=180$ (6 sec), increases the video classification accuracy while keeping it long enough to extract biological signals. Note that when we increase $\omega$\ above a certain threshold, the accuracy drops for Deep Fakes Dataset. This is due to occlusion and illumination artifacts, because the segment covers more facial variety as $\omega$\ increases. A correct balance of having sufficient segments versus sufficient length per segment is a crucial result of our analysis.

\begin{table}[h]

\centering
\begin{tabular}{c|c|c|c|c|c}\label{tab:pairw}
    $\omega$  & dataset & s. acc. & v. acc. & CNN s. & CNN v.    \\\hline
    64   & FF test & 95.75\%          & -& -& -\\
    128  & FF test & 96.55\%          & -& -& -\\
    256  & FF test & 98.19\%          & -& -& -\\
    300  & FF test & \textbf{99.39}\% & -& -& -\\\hline
    64   & FF      & 93.61\%          & -& -& -\\
    128  & FF      & 94.40\%          & -& -& -\\
    256  & FF      & 94.15\%          & -& -& -\\
    300  & FF      & \textbf{95.15\%} & -& -& -\\\hline
    128  & DF      & 75.82\%          & 78.57\% & \textbf{87.42\%} & \textbf{91.07}\%\\
    150  & DF      & 73.30\%          & 75.00\%& -& -\\
    180  & DF      & \textbf{76.78\%}          & \textbf{80.35\%}& 86.25\% & 85.71\%\\
    240  & DF      & 72.17\%          & 73.21\%& -& -\\
    300  & DF      & 69.25\%          & 66.07\%& -& -\\\hline
    128  & FF      & \textbf{77.50\%}          & \textbf{82.55}\%&\textbf{94.26\%} & \textbf{96\%}\\
    150  & FF      & 75.93\%          & 78.18\%& -& -\\
    180  & FF      & 75.87\%          & 78.85\%& 92.56\% & 93.33\%\\
    256  & FF      & 72.55\%          & 73.82\%& -& -\\
    300  & FF      & 75.00\%          & 75.16\%& -& -\\
    450  & FF      & 70.78\%          & 71.33\%& -& -\\     
    600  & FF      & 68.75\%          & 68.42\%& -& -\\     
\end{tabular}
  \caption{\textbf{Accuracy per Segment Duration.} Effects of $\omega$, on segment and video accuracies, using SVM and CNN classifiers, on FF test, entire FF, and DF datasets. First 8 rows denote pairwise task.}
  \label{tab:pairw}
\end{table}

\subsection{Face Analysis}\label{SEC_ROI}
In this section, we evaluated the dependency of our approaches on different face regions and on face detection accuracy.

\subsubsection{Size of Face Regions}
\begin{figure}[h]
\centering
  \includegraphics[width=1.05\linewidth]{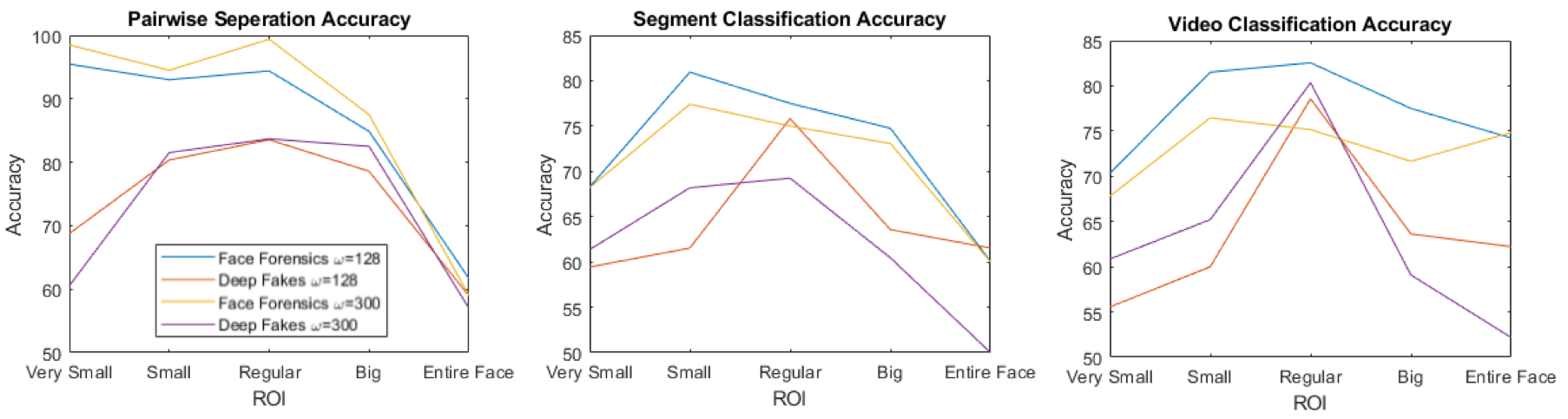}
 \caption{\textbf{ROI Comparison.} Pairwise separation (left), segment (middle), and video (right) classification accuracy per several ROIs, on FF and DF datasets.}
  \label{fig:roiplot}
\end{figure}

For our SVM classifier, biological signals are extracted from three separate regions on the face. In this section, we assess the effects of different sizes for these regions of interests. We quantize the ROIs as very small (a few pixels), small, default, big, and the whole face. Table~\ref{tab:roi} documents experiments with these ROIs, their corresponding number of pixels, dataset, segment duration ($\omega=128$ and for $\omega=300$), and number of segments, with resulting accuracies for pairwise, segment, and video tasks.  We also plot the pairwise separation, segment classification, and video classification accuracies per ROI size, on two datasets in Figure~\ref{fig:roiplot}. Lastly, we show these ROIs on a sample video frame in Figure~\ref{fig:roi}, where red contour corresponds to $G_L$ and $C_L$, green to $G_M$ and $C_M$, and blue to $G_R$ and $C_R$.  We conclude that our default ROI is a generalizable choice with a good accuracy for all cases. 

\begin{table}[h]
  \centering
\begin{tabular}{c|c|c|c|c|c|c|c}
ROI & \# pix  & data & $\omega$ & \# seg & p. acc & s. acc & v. acc
\\\hline
Smallest & 356       & FF    & 128 & 1058    & 95.46         & 68.25         & 70.33 \\
Smallest  & 356       & FF    & 300 & 400     & 98.50         & 68.18         & 67.79 \\
Smallest  & 356       & DF    & 128 & 140     & 68.76         & 59.44         & 55.55\\
Smallest  & 356       & DF    & 300 & 43      & 60.51         & 61.36         & 60.86\\
Small       & 2508      & FF    & 128 & 1058    & 93.00         & \textbf{80.94}& 81.51 \\
Small       & 2508      & FF    & 300 & 400     & 94.50         & 77.41         & 76.47 \\
Small       & 2508      & DF    & 128 & 140     & 80.35         & 61.53         & 60.00\\
Small       & 2508      & DF    & 300 & 43      & 81.53         & 68.18         & 65.21\\
Default         & 7213      & FF    & 128 & 1058    & 96.55         & 77.50         & \textbf{82.55}\\
Default         & 7213      & FF    & 300 & 400     & \textbf{99.39}& 75.00         & 75.16\\
Default         & 7213      & DF    & 128 & 140     & 83.55         & \textbf{75.82}& \textbf{78.57}\\
Default         & 7213      & DF    & 300 & 43      & \textbf{83.69}& 69.25         & 66.07\\
Big         & 10871     & FF    & 128 & 1058    & 84.87         & 74.75         & 77.50\\
Big         & 10871     & FF    & 300 & 400     & 87.50         & 73.07         & 71.66\\
Big         & 10871     & DF    & 128 & 140     & 78.58         & 63.57         & 63.63\\
Big         & 10871     & DF    & 300 & 43      & 82.51         & 60.46         & 59.09\\
Face      & 10921     & FF    & 128 & 1058    & 61.66         & 60.04         & 74.23\\
Face      & 10921     & FF    & 300 & 400     & 58.97         & 60.00         & 74.84\\
Face      & 10921     & DF    & 128 & 140     & 58.93         & 61.53         & 62.22\\
Face      & 10921     & DF    & 300 & 43      & 56.97         & 50.00         & 52.17\\

\end{tabular}
  \caption{\textbf{ROI Analysis on SVM Classifier:} Five different ROI sizes are evaluated on FF and DF datasets, with different segment sizes. Corresponding values are plotted in Figure \ref{fig:roiplot}, and corresponding ROI's are drawn on Figure \ref{fig:roi}.}
  \label{tab:roi}
\end{table}

\subsubsection{Face Detection Dependency}
Our method needs some skin to extract biological signals. We do not need a perfect face detector, however we need to find some facial area to extract the signal. Our implementation is modular enough to enable using other face detectors and handle multiple faces, both of which affects the accuracy but not significantly. In order to assess our robustness against face detection accuracy, we demonstrate results where (i) the face detection accuracy is very low, and (ii) the video has multiple faces.



\begin{figure}[h]
\centering
  \includegraphics[width=0.8\linewidth]{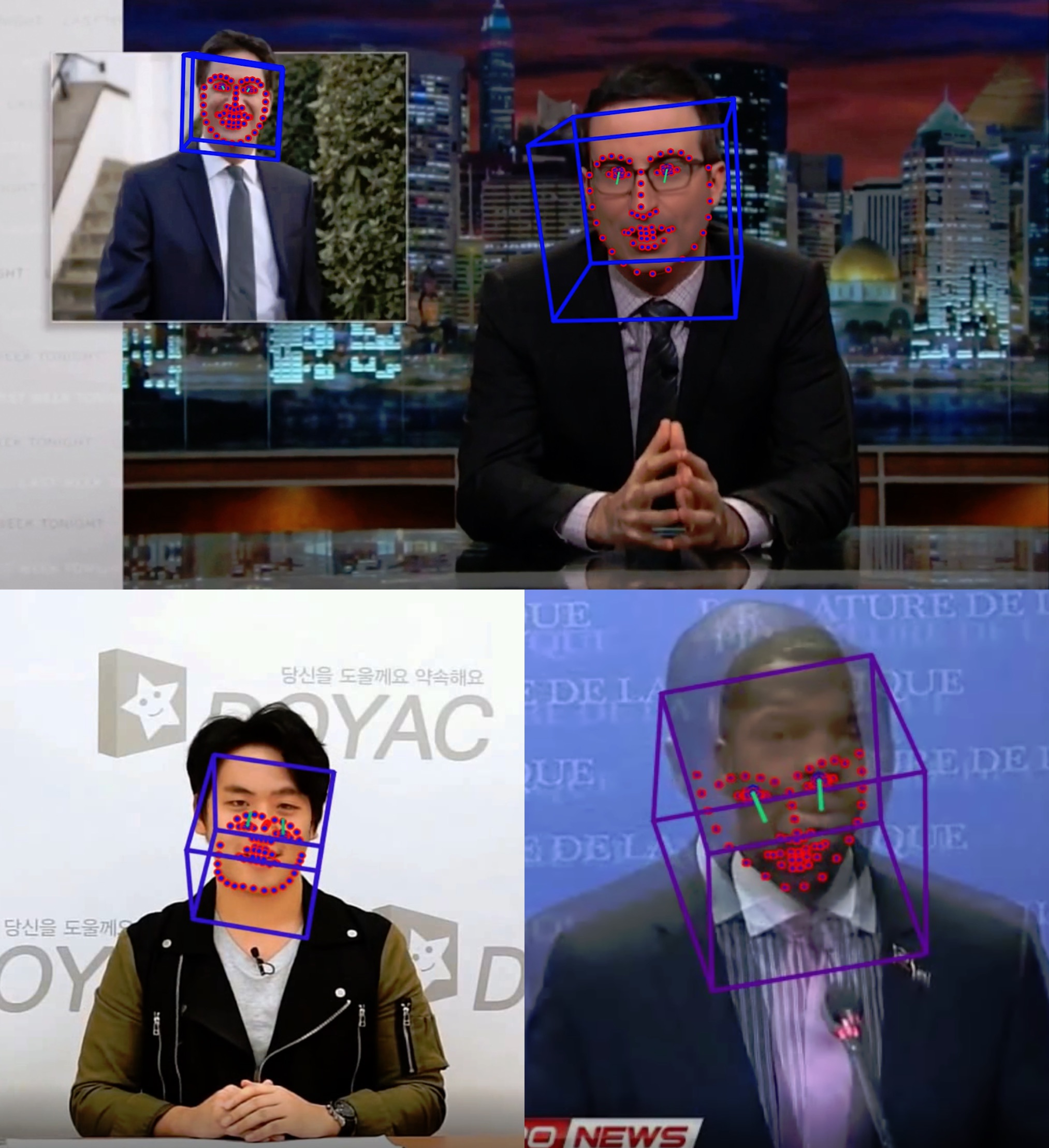}
 \caption{\textbf{Face Detection Dependency.} FakeCatcher catches a fake video with multiple faces (top) as $91.7\%$ fake, a fake video with $3\%$ face detection confidence (left) as $99.97\%$ fake, and a real video with $17\%$ face detection confidence (right) as $65.47\%$ real.}
  \label{fig:facedet2}
\end{figure}

In Figure\ref{fig:facedet2}, our results on unmodified but edge-case videos are shown. The top video contains two faces, we process them separately and we find each of them to be fake, with an average confidence of $91.7\%$. On the left is a fake video, even though its face detection accuracy is $3\%$ (very low compared to the average), FakeCatcher classifies this video as fake with $99.97\%$. Lastly, the real video on the right has a significant frame ordering problem, thus the face detector has $17\%$ accuracy. Nevertheless we classify as real with $65\%$. 

\subsection{Image Quality Analysis}
\label{SEC_Blur}
\begin{figure}[h]
\centering
  \includegraphics[width=1\linewidth]{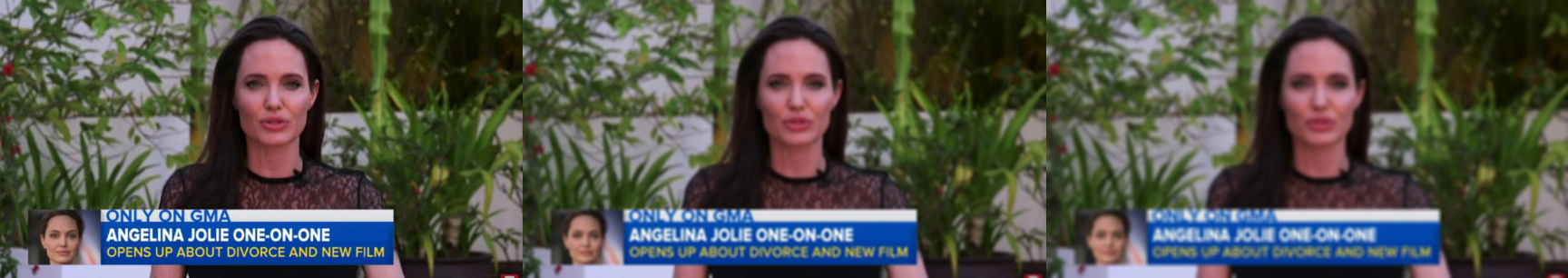}
 \caption{\textbf{Robustness.} Original (left) and blurred images with 7x7 (middle) and 11x11 (right) kernels.}
  \label{fig:gaussian}
\end{figure}

 In order to estimate an exact heart rate, the PPG signal needs to have very low SNR. In contrast, the spatio-temporal inconsistency of PPGs is sufficient for our approach to detect fakes. Even low quality authentic videos can still preserve this consistency, differentiating our dependency on low noise condition compared to the heart rate estimation task. To support this postulation, we analyze our detection accuracy with images processed with two commonly used image processing operations: Gaussian blur and median filtering. We use Celeb-DF \cite{Celeb_DF_cvpr20} dataset, with different kernel sizes, fixing $\omega=128$ segment duration (Table \ref{tab:gaussian}). The accuracy stays relatively unaffected up to a kernel size of 7x7. Then, as expected, the larger the filter kernel is used, the lower the accuracy gets. Nevertheless, such large kernel blurs modify the image significantly enough that it does not matter if the image is authentic or fake (Figure~\ref{fig:gaussian}).

\begin{table}[h]
\centering
\begin{tabular}{c|c|c}
Operation  & Kernel  & Accuracy
\\\hline
Original        & N/A     & 91.50\%\\
Gaussian blur   & 3x3     & 91.31\%\\
Gaussian blur   & 5x5     & 88.61\%\\
Gaussian blur   & 7x7     & 85.13\%\\
Gaussian blur   & 9x9     & 70.84\%\\
Gaussian blur   & 11x11   & 65.44\%\\
Median Filter   & 3x3     & 88.41\%\\
Median Filter   & 5x5     & 83.01\%\\
Median Filter   & 7x7     & 71.42\%\\
Median Filter   & 9x9     & 65.44\%\\
Median Filter   & 11x11   & 65.25\%\\
\end{tabular}
  \caption{\textbf{Robustness.} FakeCatcher accuracies on Celeb-DF dataset under different Gaussian Blur and Median filtering operations with different kernel sizes.
}
  \label{tab:gaussian}
\end{table}

\subsection{Blind Source Separation}\label{sec:blind}
\begin{figure}[h]
    \centering
    \includegraphics[width=1.1\linewidth]{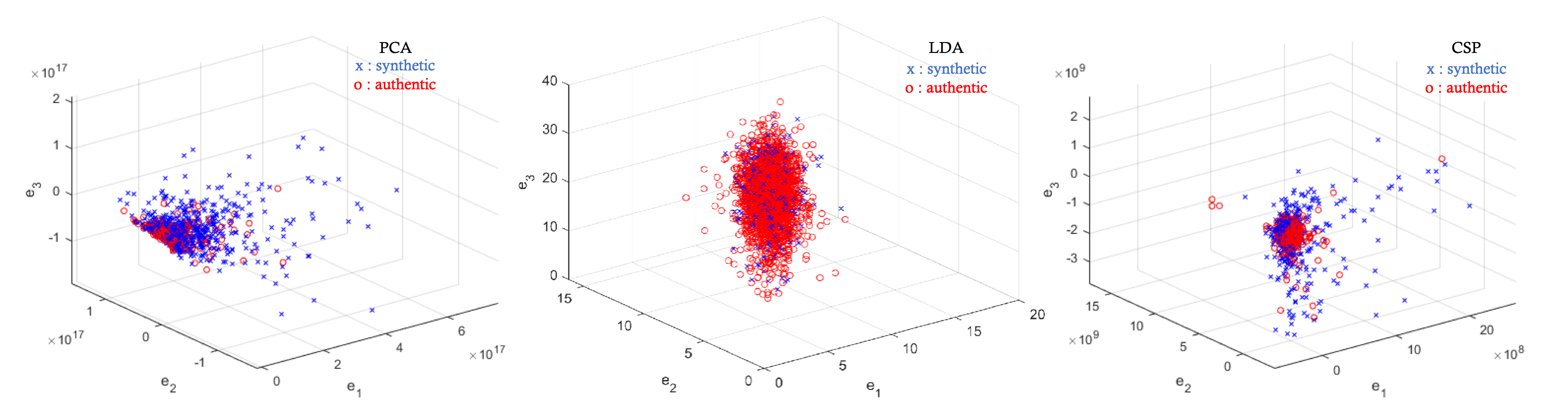}
    \caption{\textbf{Feature Space.} Original (red) and synthetic (blue) samples in three dominant component space, extracted using PCA (left), LDA(mid), and CSP (right).}
    \label{fig:pcacsp}
\end{figure}

To better understand our features, feature sets, and their relative importance, we computed the Fisher criterion detection~\cite{fishercrit} of linearly separable features if we have any. No significantly high ratio was observed, neither for LDA (linear discriminant analysis)~\cite{788121}, guiding us towards kernel based SVMs and more feature space exploration. We also applied PCA (principal component analysis)~\cite{WOLD198737} and CSP (common spatial patterns)~\cite{Koles1990} to reduce the dimensionality of our feature spaces. Figure~\ref{fig:pcacsp} shows 3D distribution of original (red) and synthetic (blue) samples by the most significant three components found by PCA, LDA, and CSP, without clear class boundaries. We also tried to condense the feature vector with our best classification accuracy. However, we achieved 71\% accuracy after PCA and 65.43\% accuracy after CSP.

\subsection{Class Accuracies}
\label{SEC_ClassBasedVideoClassification}
Throughout our paper, we use the metric as dataset accuracy in order to simplify the explanation of results. However, catching fake videos (true positives) is as important as passing real videos through (true negatives). In some defensive systems, detecting real videos as fake (false positive) is more tolerable than letting fake videos slip (false negatives), because there is a secondary manual detection process. On the other hand, too many false positives may overwhelm the system towards manually checking everything.

In order to assess our system towards these concerns, we analyzed the class accuracies of FakeCatcher in different datasets. Table\ref{tab:ClassBased} lists all dataset, real, and synthetic accuracies on all datasets used throughout the paper, and with specific generative models of FF++\cite{FF++}. All models use $\omega=128$ and a 60/40 split, except the one trained on Celeb-DF using the given subsets. We observe that there is a slight general tendency of creating false positives. We also note the special case of Celeb-DF that its class imbalance of 6.34:1 fake to real ratio bumps up this tendency. We claim that this disturbance is the side effect of the head motion, lighting, color artifacts, and other environment changes on the PPG signal. However, our probabilistic video classification, incorporating frequency domain, and appropriate segment durations negate these side effects. Following the discussion in the beginning of this section, we believe that automatic monitoring systems would prefer false positives over false negatives as long as it is not overwhelming the secondary manual process; which makes FakeCatcher a good candidate for deployment to production.

\begin{table}[h]
\centering
\begin{tabular}{c|c|c|c|c}
Dataset         & Gen. Model               & \%Dataset& \%Real& \%Synthetic
\\\hline
UADFV\cite{8683164} & FakeApp\cite{FakeApp} & 97.36\%& 94.93\%& 100\%\\ 
FF++\cite{FF++} & Face2Face\cite{f2f}            & 96.37\%  & 96.00\%& 96.75\% \\
FF\cite{ff} & Face2Face\cite{f2f} & 96\%   & 94.24\%  & 97.75\%\\
FF++\cite{FF++} & FaceSwap\cite{FaceSwap}   & 95.75\% & 94.75\%& 96.75\% \\
FF++\cite{FF++} & Deepfakes\cite{DeepFakes} & 94.87\% & 93.25\%& 96.50\%  \\
FF++\cite{FF++} & All  & 94.65\%      &88.25\%   &96.25\%\\
Celeb-DF\cite{Celeb_DF_cvpr20} &Default & 91.50\%    &76.40\%   &99.41\%\\
DF (ours) & Mixed & 91.07\% & 85.26\% & 96.89\% \\
FF++\cite{FF++} & Neural Textures\cite{10.1145/3306346.3323035} & 89.12\%& 86.75\%& 91.50\%\\

\end{tabular}
  \caption{\textbf{Class Accuracies.} Accuracies per dataset, per source, and per class, on FF, DF, FF++, and Celeb-DF.
}
  \label{tab:ClassBased}
\end{table}

\subsection{Signal Processing Enhancements}

In this subsection, we document the representativeness of several possible pre/post processing of feature sets and signals by small experiments while keeping the signal and feature set the same. 

\subsubsection{Normalization}
We evaluated the most representative pairwise segmentation result (Section \ref{sec:pairwise}) with different normalizations on the signals before computing their cross power spectral densities (Table~\ref{tab:norm}). We used the same toy FF dataset and kept $\omega=300$ constant, producing 387 segments. In this specific separation task, this analysis demonstrated that all frequencies and their ranges are needed as normalization may remove some characteristics of the signals to differentiate original and fake signals.

\begin{table}[h]
  \centering
\begin{tabular}{c|c}
Normalization  & Accuracy
\\\hline
None & \textbf{99.39} \\
2-norm & 87.34 \\
$\infty$-norm & 84.03 \\
Standardized moment & 63.55 \\
Feature scaling & 59.03 \\
Spectral whitening & 49.09 \\
Coefficient of variation & 34.03 \\
\end{tabular}
  \caption{\textbf{Normalization Effects:} Several normalization techniques are applied to the best configuration in the pairwise test.}
  \label{tab:norm}
\end{table}

\subsubsection{Band Pass Filtering}
We also analyzed if different frequency bands contributed in the generative noise. We divided the spectrum into below ($f<1.05$Hz), acceptable ($1.05$Hz$<f<3.04$Hz), and high heart rate frequency bands ($3.04$Hz$<f<4.68$Hz) as well as an additional high frequency band ($f>4.68$Hz). Table~\ref{tab:freq} documents pairwise separation accuracies of the most representative feature, for segment size $\omega=300$ on entire FF dataset (1320 segments).
\begin{table}[h]
  \centering
\begin{tabular}{c|c}
Frequency band  & Accuracy
\\\hline
0-15 &  \textbf{95.15} \\
0-1.05 &  80.58 \\
1.05-3.04 &  88.51 \\
3.04-4.68 & 88.21 \\
4.68-15 & 94.92 \\
\end{tabular}
  \caption{\textbf{Frequency Bands:} Analysis on several frequency bands for best pairwise separation accuracy on entire FF dataset.}
  \label{tab:freq}
\end{table}

\subsubsection{Frequency Quantization}\label{SEC_APPB_FREQ}
We also analyzed the effect of quantization after taking the inverse Fourier Transform of the signal. In Table~\ref{tab:bins} we verified that 256 bins were the optimum choice on the best configuration discussed in Section \ref{sec:pairwise} in the main paper, on the entire FF dataset.
\begin{table}[h]
  \centering
\begin{tabular}{c|c}\label{tab:signals2}
iFFT bins  & Accuracy
\\\hline
64 & 94.01\\
128 & 94.92\\
256 & \textbf{95.15} \\
512 & 94.92  \\
\end{tabular}
  \caption{\textbf{Quantization Evaluation:} Analysis on different bin counts for best pairwise separation accuracy on entire FF dataset.}
  \label{tab:bins}
\end{table}

\subsubsection{DCT Coefficients as Features}\label{SEC_DCT}
We experimented with using DCT of the signal ($C_M$ in this case) up to $N=\{1,2,3,4\}$ elements as a feature set, however the accuracy was surpassed as shown in the Section \ref{SEC_APPA}. The settings below in Table~\ref{tab:dct} correspond to the summation of DCT values up to $N$, respectively.

\begin{table}[h]
  \centering
\begin{tabular}{c|c||c|c||c|c}

$\omega$ & \#videos & Setting  & Accuracy &  Setting & Accuracy
\\\hline
512&62&    D&70.96   &B&77.41\\
256&236&   D&70.33   &B&72.03\\
196&323&   D&71.20   &B&77.39\\
128&532&   D&68.23   &B&72.93\\
64&1153&   D&65.04   &B&69.64\\

512&62&    C&75.80   &A&77.41\\
256&236&   C&67.18   &A&73.72\\
196&323&   C&77.39   &A&75.54\\
128&532&   C&73.30   &A&73.68\\
64&1153&   C&68.08   &A&69.55\\

\end{tabular}
  \caption{\textbf{DCT Components as Features:} Following some image spoofing papers, we evaluated accuracies with several DCT cut-offs with several segment sizes.}
  \label{tab:dct}
\end{table}

\section{Implementation Details}
For each segment, we apply Butterworth filter \cite{butter} with frequency band of $[0.7,14]$. We quantize the signal using Welch's method \cite{welch}. Then, we collect frequencies between $[h_{low},h_{high}]$, which correspond to below, in, and high ranges for heart beat. There is no clear frequency interval that accumulated generative noise, so we include all frequencies. We follow the PPG extraction methods in \cite{gppg} for ${G_L,G_M,G_R}$ and \cite{chromppg} for ${C_L,C_M,C_R}$. 

It is worth discussing that PPG signals extracted for heart rate and for our detection task are not of the same quality. For accurate heart rate estimation, PPG signal goes through significant denoising and componentization steps to fit the signal into expected ranges and periods. We observe that some signal frequencies and temporal changes that may be considered as noise for heart rate extraction actually contains valuable information in terms of fake content. For our task, we only utilize their coherence among facial regions and their consistency across segments, achieving 99.39\% pair separation accuracy on Face Forensics. Therefore, we intentionally did not follow some steps of cleaning the PPG signals with the motivation of keeping subtle generative noise. Also, we attest that even though videos undergo some transformations (e.g., illumination, resolution, and/or compression), raw PPG correlation does not change in authentic videos. 

\section{Future Work}

Our main focus in this paper is portrait videos, for which deep fakes are the most harmful. For general fake videos without humans, one possible extension of our work is to discover formulations of other spatiotemporal signals (i.e., synthetic illumination, wind) that can be faithfully extracted from original videos. 

For FakeCatcher, we see room for improvement by proposing a more complex CNN architecture. However, we want to go further and develop a ``BioGAN'' that explores the possibility of a biologically plausible generative model. Mimicking biological signals may be possible by introducing an extra discriminator whose loss incorporates our findings to preserve biological signals. This necessitates the extraction process to be approximated with a differentiable function in order to enable backpropagation. The development of BioGAN puts an expiration date on this work, however formulating a \textbf{differentiable} loss function that follows the proposed signal processing steps is not straightforward.

The dataset structure is another discussion we would like to pose for forthcoming research. For the generalizability of a detector, the data should not be biased towards known generative models. Another layer of complication which we would like to further investigate is learning from random fake and real video pairs (similar to \cite{8639163} and partially to our dataset). In our work, this conceals the actual features we want to learn, thus decreasing the accuracy for the "in the wild" case. Learning from pairs enables us to reduce the effects of other variants and makes the model focus on the generative differences, even though there is compression, motion, illumination artifacts. All pure machine learning based algorithms (as shown in Table~\ref{tab:compnn}) have a drop in accuracy compared to ours due to blindly exploiting this generative noise. On the other hand, we investigate the projection of this noise in biological signal domain, as a more descriptive interpretation, enabling our approach to outperform deeper models. 

\section{Conclusion}
In this paper, we present FakeCatcher, a fake portrait video detector based on biological signals. We experimentally validate that spatial coherence and temporal consistency of such signals are not well preserved in GAN-erated content. Following our statistical analysis, we are able to create a robust synthetic video classifier based on physiological changes. Furthermore, we encapsulate those signals in novel PPG maps to allow developing a CNN-based classifier, which further improves our accuracy and is agnostic to any generative model. We evaluate our approach for pairwise separation and authenticity classification, of segments and videos, on Face Forensics \cite{ff} and newly introduced Deep Fakes Dataset, achieving 99.39\% pairwise separation accuracy, 96\% constrained video classification accuracy, and 91.07\% in the wild video classification accuracy. These results also verify that FakeCatcher detects fake content with high accuracy, independent of the generator, content, resolution, and quality of the video.

Apart from the FakeCatcher and the Deep Fakes Dataset, we believe that a main contribution of this paper is to provide an in-depth analysis of deep fakes in the wild. To our knowledge, generative models are not explored by biological signals before, and we present the first experimental study for understanding and explaining human signals in synthetic portrait videos. We hope that our findings will illuminate the future research in defense against deep fakes. Lastly, we encourage continuation of generalizable fake detection research by making our Deep Fakes Dataset available to the research community.

\ifCLASSOPTIONcompsoc
  \section*{Acknowledgments}
\else
  \section*{Acknowledgment}
\fi
We would like to thank Prof. Daniel Aliaga for proofreading an earlier version of the manuscript. We would also like to thank for the supportive research environments in the authors' current and previous institutions, and funding resources as NSF (CNS-1629898), Center of
Imaging, Acoustics, and Perception Science, and the Research Foundation of Binghamton University. Lastly, we would like to acknowledge the reviewers' constructive feedback towards improving the experimental evaluation and the clarity of our approach.

\ifCLASSOPTIONcaptionsoff
  \newpage
\fi



%
\bibliographystyle{IEEEtran}
\bibliography{IEEEabrv,fakecatcher}

%


\begin{IEEEbiography}[{\includegraphics[width=1in,height=1.25in,clip,keepaspectratio]{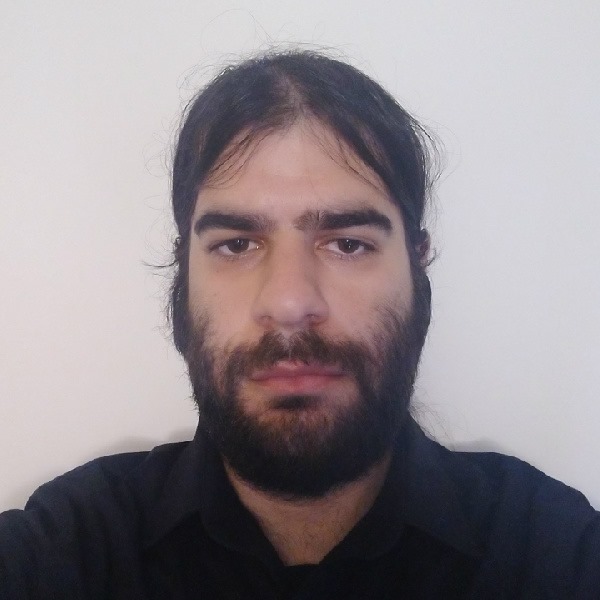}}]{Umur Aybars Ciftci}
received his MSc degree in computer science from Binghamton University in 2014. He is currently a PhD candidate in the computer science department of Binghamton University where he is a member of Graphics and Image Computing Laboratory. His research interests are in computer vision, human computer interaction, and affective computing.  
\end{IEEEbiography}

\begin{IEEEbiography}[{\includegraphics[width=1in,height=1.25in,clip,keepaspectratio]{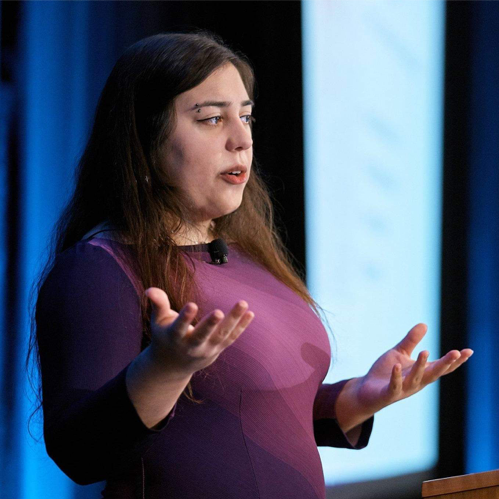}}]
{Ilke Demir}
earned her PhD degree in Computer Science from Purdue University, focusing on 3D vision approaches for generative models, urban reconstruction and modeling, and computational geometry for synthesis and fabrication. Afterwards, Dr. Demir joined Facebook as a Postdoctoral Research Scientist working with Ramesh Raskar from MIT. Her research included human behavior analysis via deep learning in virtual reality, geospatial machine learning, and 3D reconstruction at scale. In addition to her publications in top-tier venues (SIGGRAPH, ICCV, CVPR), she has organized workshops, competitions, and courses in the intersection of deep learning and computer vision. She has received several awards and honors such as Jack Dangermond Award, Bilsland Dissertation Fellowship, and Industry Distinguished Lecturer, in addition to her best paper/poster/reviewer awards. Currently she leads the research efforts on 3D vision and deep learning approaches in the world's largest volumetric capture stage at Intel Studios.
\end{IEEEbiography}


\begin{IEEEbiography}[{\includegraphics[width=1in,height=1.25in,clip,keepaspectratio]{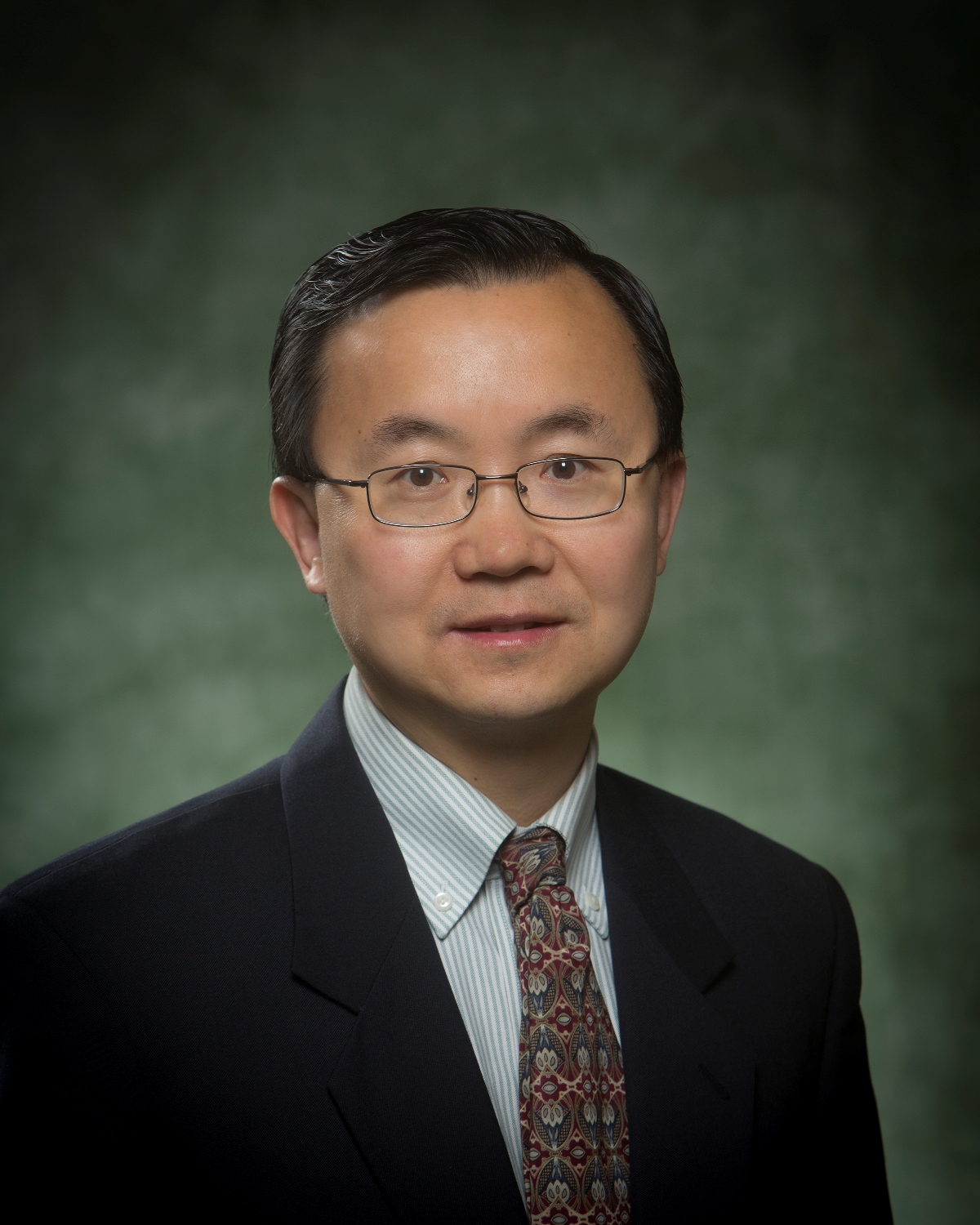}}]{Lijun Yin}
Lijun Yin is a Professor of Computer Science, Director of Center for Imaging, Acoustics, and Perception Science at Binghamton University, Director of Graphics and Image Computing Laboratory, and Co-director of Seymour Kunis Media Core, T. J Watson School of Engineering and Applied Science at the State University of New York at Binghamton.  He received Ph.D. of computer science from the University of Alberta and Master of Electrical Engineering from Shanghai Jiao Tong University. Dr. Yin`s research focuses on computer vision, graphics, HCI, and multimedia, specifically on face and gesture modeling, analysis, recognition, animation, and expression understanding. His research has been funded by the NSF, AFRL/AFOSR, NYSTAR, and SUNY Health Network of Excellence. Dr. Yin received the prestigious Lois B. DeFleur Faculty Prize for Academic Achievement Award (2019), James Watson Investigator Award of NYSTAR (2006), and SUNY Chancellor's Award for Excellence in Scholarship \& Creative Activities (2014). He holds 11 US patents, and released four 2D/3D/4D facial expression databases to public, and has published over 150 papers in technical conferences and journals. Dr. Yin served as a program co-chair of FG 2013 and FG2018. He is currently serving on editorial board of IVC and PRL. 
\end{IEEEbiography}




\end{document}